\documentclass[a4paper,fleqn]{cas-dc}
\usepackage{float}     
\usepackage{placeins}  

\usepackage[authoryear,round]{natbib}
\usepackage{enumitem}
\DeclareUnicodeCharacter{202F}{~}
\usepackage{subcaption}
\usepackage{tikz}
\usepackage{etoolbox}

\makeatletter
\@ifundefined{printorcid}{}{%
  \renewcommand{\printorcid}[1]{}}   
\@ifundefined{orcidlink}{}{%
  \renewcommand{\orcidlink}[1]{}}   
\@ifundefined{orcidicon}{}{%
  }     
\makeatother

\begin{document}

\let\WriteBookmarks\relax
\def\floatpagepagefraction{1}
\def\textpagefraction{.001}

\shorttitle{Targetless Registration}    

\shortauthors{Marco Antonio Ortiz Rincón}  

\title [mode = title]{Quality-controlled registration of urban MLS point clouds reducing drift effects by adaptive fragmentation}

\tnotemark[1]

\author{Marco Antonio Ortiz Rincón}
\cormark[1]

\ead{marco.ortiz@tum.de}

\affiliation{organization={Chair of Engineering Geodesy, TUM School of Engineering and Design, Technical University of Munich},
            addressline={Arcisstr. 21}, 
            postcode={80333}, 
            city={Munich},
            country={Germany}}

\author{Yihui Yang}

\ead{yihui.yang@tum.de}

\author{Christoph Holst}

\cortext[1]{Corresponding author}

\begin{abstract}
This study presents a novel workflow designed to efficiently and accurately register large-scale mobile laser scanning (\textit{MLS}) point clouds to a target model point cloud in urban street scenarios. This workflow specifically targets the complexities inherent in urban environments and adeptly addresses the challenges of integrating point clouds that vary in density, noise characteristics, and occlusion scenarios, which are common in bustling city centers. Two methodological advancements are introduced. First, the proposed Semi-sphere Check (\textit{SSC}) preprocessing technique optimally fragments \textit{MLS} trajectory data by identifying mutually orthogonal planar surfaces. This step reduces the impact of \textit{MLS} drift on the accuracy of the entire point cloud registration, while ensuring sufficient geometric features within each fragment to avoid local minima.
Second, we propose Planar Voxel-based Generalized Iterative Closest Point (\textit{PV-GICP}), a fine registration method that selectively utilizes planar surfaces within voxel partitions. This pre-process strategy not only improves registration accuracy but also reduces computation time by more than 50\% compared to conventional point-to-plane \textit{ICP} methods.
Experiments on real-world datasets from Munich’s inner city demonstrate that our workflow achieves sub-0.01 m average registration accuracy while significantly shortening processing times. The results underscore the potential of the proposed methods to advance automated 3D urban modeling and updating, with direct applications in urban planning, infrastructure management, and dynamic city monitoring.

\end{abstract}






\begin{keywords}
point cloud registration
\sep mobile laser scanning
\sep fragmentation
\sep voxels
\sep 3D urban model updating
\end{keywords}

\maketitle



\section{Introduction}\label{Introduction}
Urban environments are becoming increasingly dynamic, characterized by rapid structural changes that necessitate frequent updates to 3D city models \citep{ArroyoOhori2018}. Maintaining up-to-date models is essential for a range of applications, including urban planning, infrastructure management, and the monitoring of urban development \citep{Eriksson2021}. However, traditional update methods — often based on manual inspection and editing — are labor-intensive and susceptible to human errors \citep{Wang2020}. As a result, 3D point cloud data acquired through laser scanning, particularly Mobile Laser Scanning (\textit{MLS}), has emerged as a powerful alternative due to its ability to provide detailed, accurate, and efficient 3D spatial information \citep{Wang2020}.

The point cloud can be regarded as a virtual representation of the real world. These kinds of 3D data sets typically consist of two groups: source and target point clouds. Target point clouds are those that have been cleaned up and accurately georeferenced (e.g., 3D data provided by the urban geo-data service). In contrast, the source point clouds are those that need to be processed and aligned with the target point cloud (e.g., acquired \textit{MLS point clouds}  ). These point clouds may contain thousands to billions of points, enabling the representation and update of the surface details of the urban environment.

Despite their advantages, integrating diverse point cloud datasets remains challenging due to their varying density, measurement noise, and low overlaps in different data sources \citep{Liu2022}. Conventional point cloud registration techniques typically struggle with efficiency and accuracy when applied to large-scale urban datasets \citep{Lee2024}. Standard \textit{ICP}-based fine registration is highly time-consuming for city-scale point clouds. Furthermore, a single transformation matrix may not be enough to achieve an accurate alignment for large-scale point clouds due to potential non-rigid areas. For instance, point clouds captured by the \textit{MLS} system may drift over time. To address this issue, \textit{MLS} data can be divided into smaller fragments with either equal time intervals or equal length and registered to the reference data individually \citep{lucks2021improving, lucksRaffl, zhu2020robust, xu2024precisereg}. However, these approaches may fail in certain sections that lack distinct features or contain a high level of noise within a small fragment \citep{FrameFail}. On the other hand, the non-rigid effect can still occur if the fragments are too large.

To overcome these limitations, this study develops a comprehensive pipeline to accurately and efficiently register \textit{MLS} point clouds to the existing reference point cloud data in the context of city street scenarios. To tackle the challenges in conventional point cloud registration workflows, two novel modules are proposed in this workflow. Additionally, a drift analysis strategy is employed to correct errors in the source point cloud:

\begin{itemize}
    \item \textbf{Semi-sphere check fragmentation (\textit{SSC})} – an adaptive \textit{MLS} point cloud fragmentation strategy using normal-based validation in a semi-sphere space for each \textit{MLS} fragment. This module is able to generate high-quality fragments containing sufficient mutually orthogonal planes, thus reducing the drift influence as well as keeping enough features for robust registration.
    \item \textbf{Planar voxel-based Generalized-\textit{ICP} (\textit{PV-GICP})} – a fast fine registration method that automatically selects planar voxels to perform Generalized-\textit{ICP} (\textit{GICP}), achieving high accuracy while significantly reducing runtime.
    \item \textbf{Drift analysis} - a method to identify and reduce drift errors through fragmentation processes for evaluating and enhancing the accuracy of \textit{MLS} source point clouds.
    
\end{itemize}

The proposed pipeline is applied and evaluated on a large-scale dataset from the inner-city area of Munich and evaluated by quantifying the Multiscale Model-to-Model Cloud Comparison (\textit{M3C2}) distances on selected stable patches of registered scans \citep{M3C2}. Besides, the drift effects in \textit{MLS} are effectively quantified and analyzed by the derived transformation parameters of each fragment. Addressing critical gaps in existing registration methods, the proposed workflow exhibits high potential in automated 3D urban model updates. 

The rest of this article is organized as follows: The next section addresses the related work in current point cloud registration techniques. Section~\ref{Methodology} presents the principles and details of the proposed registration pipeline. Section~\ref{ExperimentalEvaluation} demonstrates the experimental results on a street dataset with three different scenarios. Section~\ref{Discussion} discusses the limitations and potential improvements of the method, followed by conclusions in Section~\ref{Conclusion}.


\section{Related work}\label{Related work}
Laser scanning technology has become an efficient tool for capturing detailed 3D point cloud data, enabling accurate and dynamic urban model creation \citep{Wang2020}. \textit{MLS} technology is particularly widely used for street point cloud data acquisition due to its convenience, efficiency, and lower cost compared to aerial laser scanning (\textit{ALS}) and terrestrial laser scanning (\textit{TLS}). Regarding \textit{MLS} point cloud registration tasks, we divide the related work into \textit{MLS} point cloud fragmentation (Section~\ref{Preprocess}), coarse registration (Section~\ref{CoarseRegistration}), and fine registration (Section~\ref{FineRegistration})

\subsection{\textit{MLS} point cloud fragmentation}\label{Preprocess}
For registration tasks involving \textit{MLS} point clouds, effective preprocessing is crucial to reduce noise, improve computational efficiency, and enhance registration accuracy. When dealing with \textit{MLS} data that has long trajectories, an important preprocessing step is dividing the \textit{MLS} point clouds into appropriate segments that contain sufficient geometric features.

Typically, \textit{MLS} data fragmentation is carried out using equispatial or equitemporal splits \citep{lucks2021improving, zhu2020robust, xu2024precisereg, XU2025PL4U}. More recent approaches utilize geometric analyses, including normal vector clustering \citep{jiang2017nvcluster} and curvature-based segmentation \citep{lee2021curvseg}. In some cases, when \textit{ALS} is used, a strip adjustment pipeline can be implemented. In this framework, one complete flight strip, a single laser line flown from take-off to landing, serves as the basis for processing \citep{sun2023stripadjust}. 

For indoor scenarios, \citet{mahmood2020crosssec} samples each indoor \textit{MLS}/\textit{TLS} scan (as well as the IFC-derived BIM) at $0.10m$ height intervals, resulting in a stack of horizontal cross-sections. They select the slice with the maximal convex-hull area, an empirical indicator of “least occlusion and richest geometry”, to treat that 2-D slice as a single rigid fragment for subsequent line-based coarse-to-fine registration. Additionally,  \citet{koszyk2024segment} first applies PointNet++-based semantic segmentation to both \textit{MLS} and \textit{UAV} point clouds, then tessellates each cloud into square sub-regions. Only the tiles that contain points from both datasets are treated as rigid fragments. Their centroids and local density cues are then used to initiate a coarse tile-to-tile match before fine \textit{ICP} refinement.

Nonetheless, these methods generally do not ensure that each fragment contains sufficiently distinguished geometric features, particularly regarding mutual surface orientations, while also maintaining a limited size.

\subsection{Coarse registration}\label{CoarseRegistration}
\textit{MLS} point clouds that have not been georeferenced typically differ significantly from the reference data, necessitating effective coarse registration prior to fine registration. Feature-based matching remains the dominant paradigm for initial alignment: Most methods establish corresponding features along with RANSAC to estimate an optimal rigid transformation \citep{fischler1981ransac,rusu2009fpfh}. Global-set matching methods such as 4PCS \citep{aiger2008fourpcs} and Super4PCS \citep{mellado2014super4pcs} accelerate wide baseline alignment by exploiting congruent four-point bases. Graph-theoretic filters like \textit{GROR} \citep{yan2022gror} and certifiable optimizers such as TEASER++ \citep{yang2020teaserpp} further prune outliers prior to pose estimation. Learning-based frameworks, including PointNetLK \citep{aoki2019pointnetlk} and Deep Global Registration (DGR) \citep{choy2020dgr}, directly regress the 6-DOF pose from raw points, but require substantial training data and are sensitive to domain shifts.

\subsection{Fine registration}
\label{FineRegistration}

Standard Iterative Closest Point (\textit{ICP}) \citep{besl1992icp} method minimizes the point-to-point error, while point-to-plane \textit{ICP} improves convergence speed and reduces the impact of non-uniform point densities \citep{PointtoPlane}. Generalized \textit{ICP} \citep{segal2009gicp} unifies the two viewpoints, and Go-\textit{ICP} \citep{yang2015goicp} offers a globally optimal branch-and-bound solution at significant computational cost. Probabilistic variants, including the Normal Distributions Transform \citep{biber2003ndt,magnusson2009ndt} and Coherent Point Drift \citep{myronenko2010cpd}, replace nearest-neighbor search with distribution fitting for added robustness. More recent advancements exploit planar patches in identified stable areas to perform point-to-plane \textit{ICP} \citep{YangSchwiegerSVreg, yang2025piecewise} or using adaptive robust kernels to mitigate outliers \citep{bouguelia2018robustkernel}. 

Despite these advances, city-scale \textit{MLS} data registration still suffers from long processing times and sensitivity to non-planar and highly rough structures. Our methodology, introduced in the next section, overcomes those limitations.

\section{Methodology}\label{Methodology}

\begin{figure*}
	\centering
	\includegraphics[width=1\textwidth]{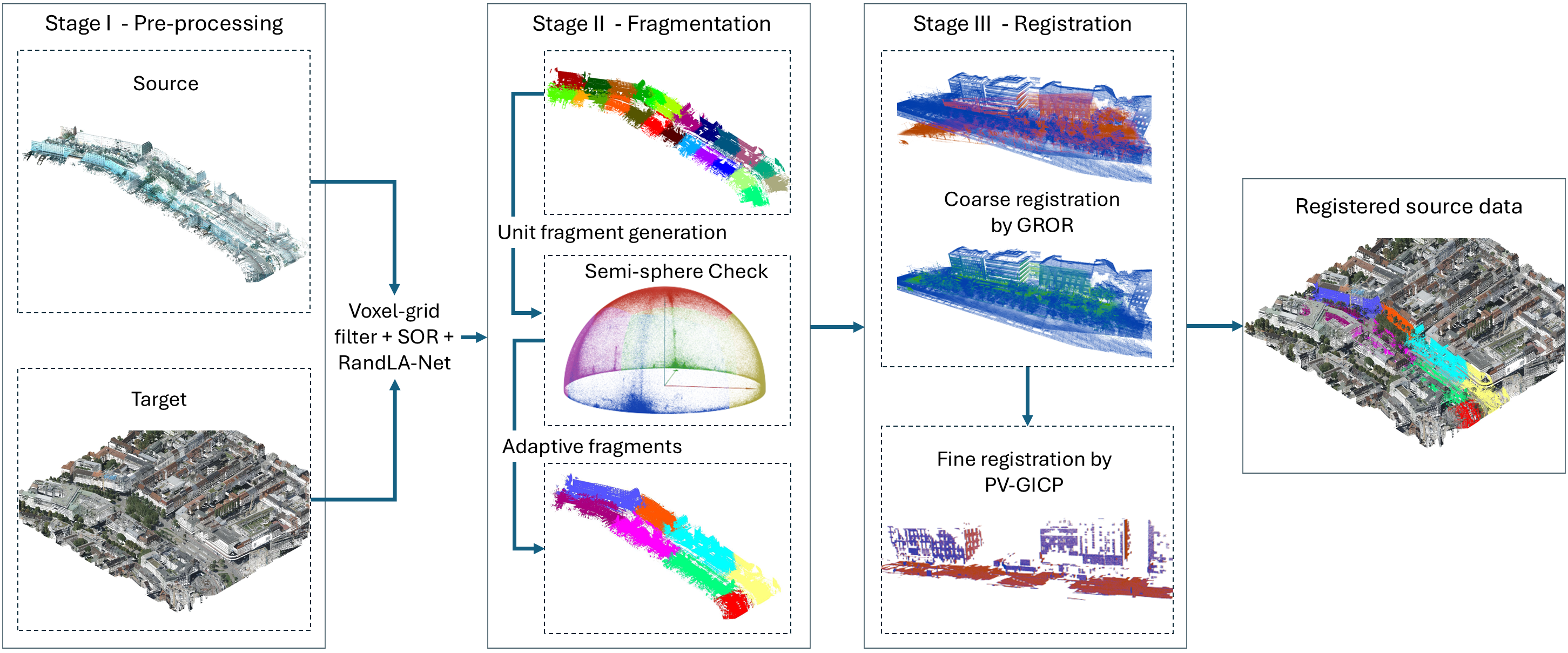}
	\caption{Flowchart illustrating the proposed targetless registration workflow, integrating the novel Semi-sphere Check fragmentation (\textit{SSC}) and planar voxel-based Generalized \textit{ICP} (\textit{PV-GICP}) techniques.}
	\label{FIG:1}
\end{figure*}

To solve the aforementioned challenges of \textit{MLS} point cloud registration in urban street scenarios, we propose a holistic workflow, as shown in Figure \ref{FIG:1}. This pipeline includes data pre-processing, data fragmentation, and coarse-to-fine registration, as further explained in the next sub-sections.
As input of our workflow, we take the source and the target point clouds. Those might already have been pre-processed by the stakeholders providing the data (surveying companies performing the update measurements, manufacturers improving their point cloud quality, authorities maintaining the target point clouds, etc.). The workflow is designed to operate independently of instrument-specific properties.

\subsection{Stage I - Data pre-processing}\label{sec:data-prep}

Raw \textit{MLS} point clouds are first resampled to a uniform resolution of
0.02 m using a voxel-grid filter to improve the following data processing efficiency ~\citep{rusu2011pcl}. Outliers are removed in two steps. First, a Statistical Outlier Removal (\textit{SOR}) is employed to eliminate isolated noise. Remaining points are then semantically classified by RandLA-Net using a pre-trained model~\citep{RandLA}. Subsequently, dynamic objects (e.g., pedestrians, vehicles, foliage, etc.) are discarded, leaving solely static parts that include buildings and ground. This step reduces the size of the point cloud by more than 30\% while preserving all stable structures relevant for registration.

\subsection{Stage II - Fragmentation using Semi-sphere Check}
\label{sec:semisphere}

\textit{MLS} data can provide not only 3D point clouds of the scanned scene but also trajectory information with time information. Large-scale \textit{MLS} point clouds in urban street environments can thereby be divided into several small fragments according to the given temporal information or trajectory length. Based on the \textit{GPS} time of each measurement, the scan is empirically divided into consecutive 10 s intervals as initial fragments. The number of fragments, thus, scales with the overall point cloud length. This temporal fragmentation can effectively reduce the drift effect during scanning \citep{XU2025PL4U}. For \textit{MLS} data without available \textit{GPS} time, the point cloud can be spatially fragmented into equal-sized sections with a 10m trajectory. 

After generating initial fragments with either equal temporal or spatial intervals, Semi-sphere Check (\textit{SSC}) is applied to decide whether each fragment contains sufficient geometric features for reliable registration.  The key idea of \textit{SSC} is to merely retain fragments that contain surfaces in all three orthogonal directions, particularly for building façades in urban scenes. In detail, it contains the three steps as follows.

\paragraph{Step 1: Distance validation.}
The travelled distance of \textit{MLS} depends on the scanning time and the platform's speed. When the platform stops (e.g., due to a traffic light),   the \textit{GPS} time keeps advancing, thus capturing a limited scene within a fixed 10 s interval. To avoid generating fragments with too little spatial extent, the trajectory length of each initial fragment is computed. If the associated trajectory length is less than 10 m, the subsequent initial fragment with a 10 s interval will be appended to form a new fragment.

\begin{figure}
  \centering

  \begin{subfigure}[a]{0.8\columnwidth}
    \centering
    \includegraphics[width=\textwidth]{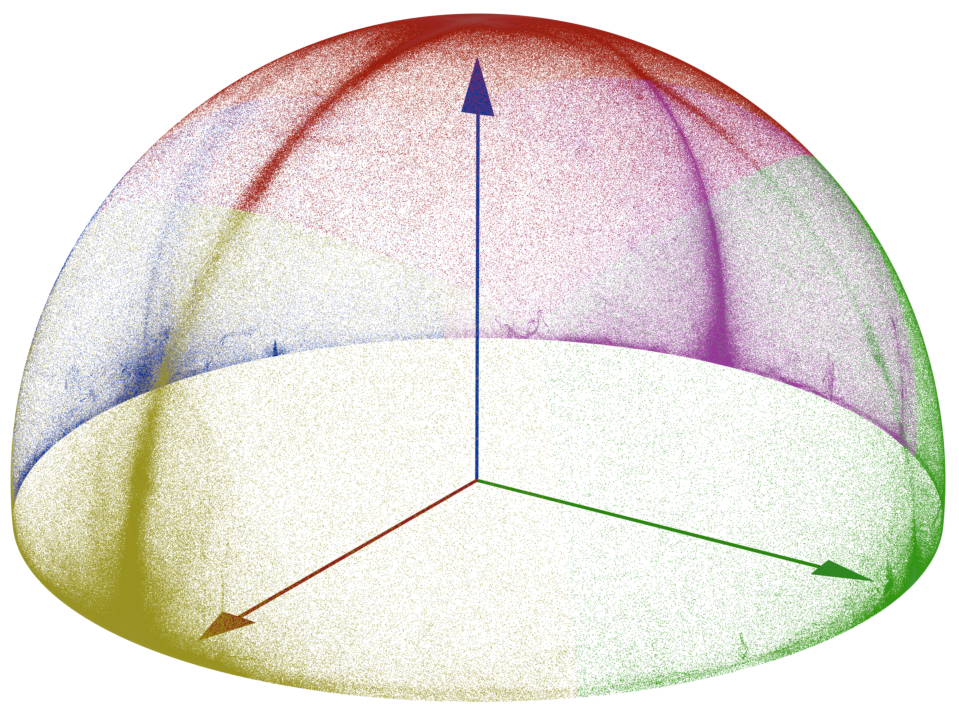}  
    \caption{Normal distribution on a Semi Sphere}
    \label{fig:frames-a}
  \end{subfigure}

  \vspace{4pt}  

  \begin{subfigure}[b]{0.8\columnwidth}
    \centering
    \includegraphics[width=\textwidth]{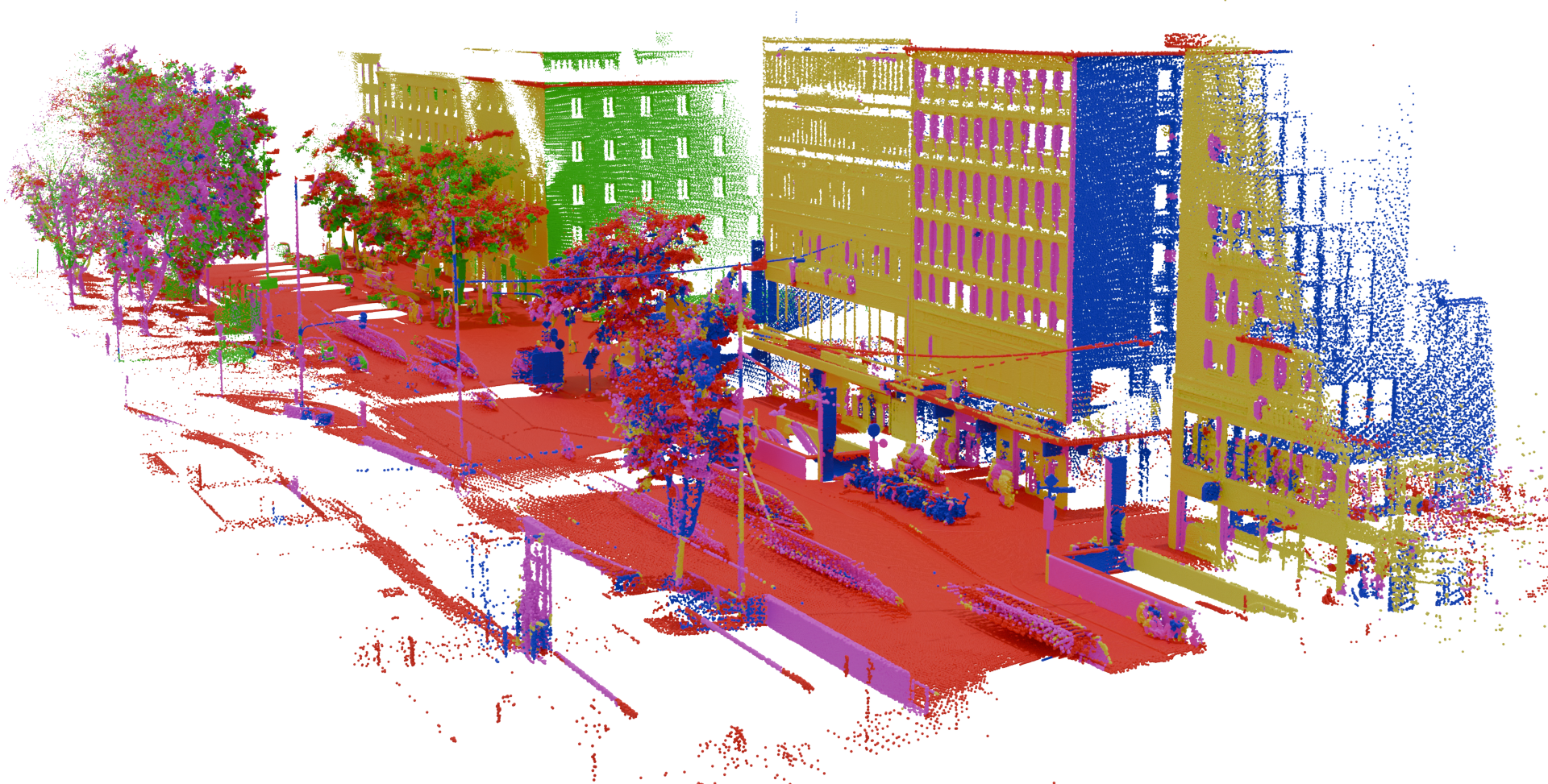} 
    \caption{Validated frame after the \textit{SSC}}
    \label{fig:frames-b}
  \end{subfigure}

  \caption{(a) Projecting all normal vectors of a fragment onto a semi-spherical domain to facilitate the distribution analysis. (b) Example of a fragment of a street \textit{MLS} data which is successfully validated by the \textit{SSC}, indicating adequate coverage of mutually orthogonal planar surfaces.}
  \label{fig:semisphere-frames}
\end{figure}

\paragraph{Step 2: Normal‐vector analysis in a unit semi-sphere.}
For every validated fragment in Step 1, we compute surface normals after aligning the longest bounding‐box axis with the global~\(x\)-axis. The normals are then oriented towards the fragment centroid to ensure a consistent outward direction. All oriented normals are projected onto the surface of a unit semi-sphere, producing a spatial distribution as illustrated in Figure~\ref{fig:frames-a}. Since only the absolute value of the \(z\)-component is retained, the lower half‐sphere is mirrored onto the upper one.

\paragraph{Step 3: K-means clustering.}
K-means algorithm is adopted and initialized by five canonical axes
\((\pm1,0,0),\,(0,\pm1,0),\\(0,0,1)\) as seed points (Figure~\ref{fig:3d_coordinate_system}). After achieving convergence of K-means clustering, the dispersion of the seed points in the semi-sphere is computed. Their average displacement from the origin and the standard deviation are then computed. If the displacement exceeds a user-defined threshold, indicating an uneven normal distribution, a subsequent initial fragment is appended. This \textit{SSC} process is conducted iteratively and will stop when the average displacement of seed points is less than the threshold or a maximum size of the updated fragment is reached. An example of a validated fragment by this method is shown in Figure~\ref{fig:frames-b}, where five clusters are clearly represented.

\definecolor{mediumgreen}{rgb}{0,0.7,0}

\begin{figure}
    \centering
    \begin{tikzpicture}[scale=3,>=stealth, line width=1pt]

        \draw[->, mediumgreen, line width=1pt] (0,0) -- (1,0) node[anchor=north] {$(0,1,0)$}; 
        \draw[-, mediumgreen, line width=1pt] (0,0) -- (-1,0) node[anchor=north] {$(0,-1,0)$}; 

        \draw[-, red, line width=1pt] (0,0) -- (0,0,-1) 
            node[anchor=east, xshift=40pt] {$(-1,0,0)$}; 
        \draw[->, red, line width=1pt] (0,0) -- (0,0,1) 
            node[anchor=east] {$(1,0,0)$};  

        \draw[->, blue, line width=1pt] (0,0) -- (0,1) 
            node[anchor=south] {$(0,0,1)$}; 

    \end{tikzpicture}
    \caption{Illustration of initial K-means clustering seed points, positioned along canonical coordinate axes to guide the \textit{SSC} evaluation of normal-vector distributions.}
    \label{fig:3d_coordinate_system}
\end{figure}
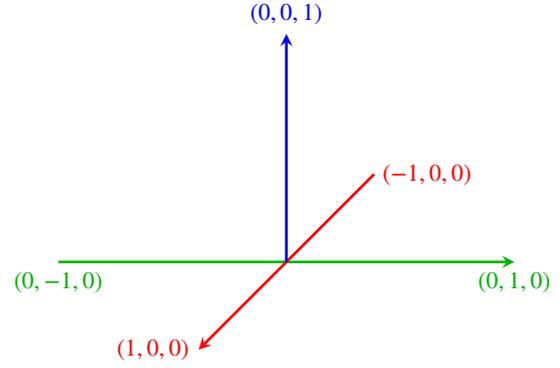

The generated fragments that exhibit sufficient orientations in three orthogonal directions are individually forwarded to the registration stage. This fragmentation strategy can effectively mitigate the drift effects in \textit{MLS} and ensure that the entire \textit{MLS} point cloud with a long trajectory is optimally registered to the reference point cloud, as described in Section \ref{Drift analysis}.

\subsection{Stage IIIa - Coarse registration}\label{sec:coarse}

After fragmentation, each fragment of the \textit{MLS} point cloud, along with the reference point cloud, is fed to the coarse registration module. Herein, Intrinsic Shape Signatures (\textit{ISS}) \citep{ISS} is used to locate geometric keypoints that are further described by the Fast Point Feature Histogram (\textit{FPFH}) \citep{rusu2009fpfh}. These keypoints are matched through a simple nearest-neighbour search in the feature space, which yields an initial correspondence set that may contain a considerable number of outliers in the context of urban street scenarios. Therefore, an automatic outlier-rejection strategy is necessary to remove the incorrect correspondences.

To enhance both robustness and computational efficiency in correspondence filtering, we adopt the Graph-Reliability Outlier Removal (\textit{GROR}) method \citep{yan2022gror}, which is specifically designed to eliminate spurious matches from an initial set of correspondences. \textit{GROR} formulates the problem as a fully connected graph, where each correspondence is treated as a node. The edge weights are defined by the Euclidean distances between the 3D endpoints of correspondence pairs, thereby capturing their spatial consistency. To evaluate the reliability of each correspondence, \textit{GROR} computes a global consistency score using the row sum of the adjacency matrix. This score reflects how well each node (i.e., match) aligns with others in terms of geometric agreement.
By selecting the top \(K\) correspondences with the highest consistency scores (e.g., with \(K\) = 800 empirically chosen for each fragment in urban scenes), the method effectively suppresses outliers and significantly improves the reliability of the input for subsequent transformation estimation.

\subsection{Stage IIIb - Fine registration}\label{sec:fine}
In street scenarios examined in our study, elevated noise levels are consistently encountered. Given that the keypoints extracted for coarse registration can be both limited and noisy. Further refinement of the coarse registration results is necessary. To address the shortcomings in accuracy and efficiency of standard \textit{ICP} in large-scale \textit{MLS} point cloud registration, we develop a fast fine registration method in this paper, which employs Generalized \textit{ICP} solely on validated stable and planar voxels from both point clouds.
This Planar voxel-based Generalized \textit{ICP} (\textit{PV-GICP}) method effectively utilizes planar structures (such as buildings and roads) in urban environments, and it improves registration robustness while significantly reducing computation time, as shown in Section \ref{Results of fine registration}.

\begin{figure*}
	\centering
	\includegraphics[width=1\textwidth]{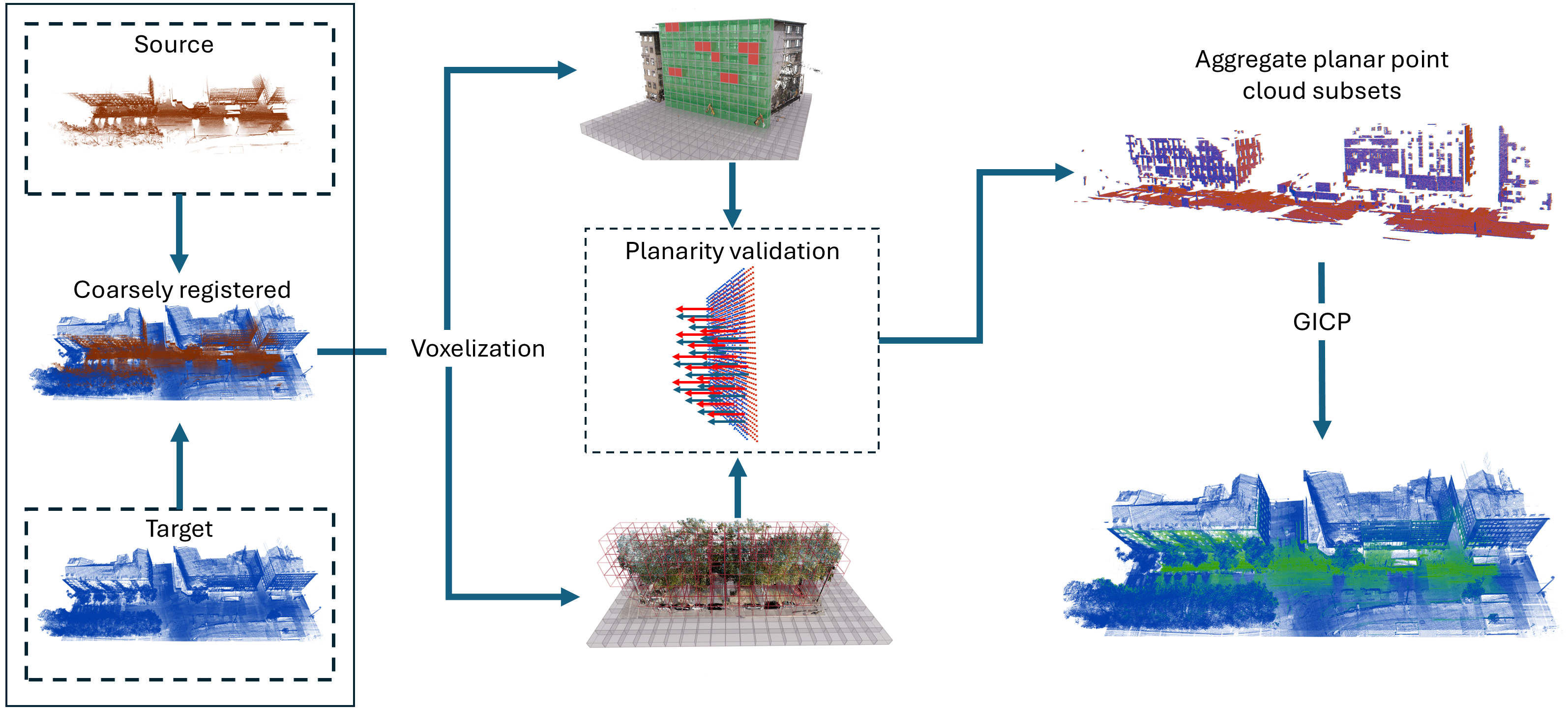}
	\caption{Workflow of \textit{PV-GICP} consisting of the selection of planar voxels and Generalized \textit{ICP}.}
	\label{FIG:PV-GICP}
\end{figure*}

In \textit{PV-GICP}, the axis-aligned bounding box (\textit{AABB}) of each cloud is first computed and merged into a single \textit{AABB} that encloses both clouds. The fused cloud is then partitioned into equal-sized voxels with a user-specified edge length.

For each voxel with sufficient points (e.g., \(\ge100\)), the normal vector of each point is estimated. The union of all normals
\(\{\,\mathbf{n}_1,\dots,\mathbf{n}_m\}\) (\(m\) is the number of all points in two corresponding voxels) is averaged, thus their mean direction is computed by

\begin{equation}
    \bar{\mathbf{n}}=\frac{1}{m}\sum_{i=1}^{m}\mathbf{n}_i,\qquad
\lVert\bar{\mathbf{n}}\rVert\neq0
\end{equation}

The normal \(\mathbf{n}_i\) of which the angle between \(\mathbf{n}_i\) and \(\bar{\mathbf{n}}\) is less than a threshold (e.g., $10^{\circ}$ is used herein)) is regarded as consistent.  Let \(C\) be the ratio of consistent normals in the voxel. If \(C\) is larger than 70\%, the corresponding voxels are deemed planar and their points are appended to two aggregated planar
clouds (source and target). Thereby, voxels that fail the planarity validation are discarded, leaving fewer but reliable points for the following fine registration.

Afterward, \textit{GICP} is performed on the two planar point clouds to estimate the rigid transformation. The final transformation parameters between each fragment and the reference point cloud can be obtained by multiplying the transformation matrices computed from coarse registration and fine registration.

\section{Experimental evaluation}\label{ExperimentalEvaluation}

The proposed workflow is applied to a \textit{LiDAR} point cloud dataset that digitized a street scene in the city of Munich. This dataset includes a reference point cloud and an updated Mobile Laser Scanning (\textit{MLS}) point cloud, which represents a new, current state of the same street taken at a different time. This will help demonstrate and evaluate the performance of the workflow. The registration accuracy and computational efficiency are systematically evaluated. Besides, based on the derived transformation parameters of each fragment, the drift effects in the \textit{MLS} process are also quantitatively analyzed.

\subsection{Data description}
\label{Datadescription}
The point cloud dataset was provided by the City of Munich, GeodatenService \footnote{\url{http://www.geodatenservice-muenchen.de}}. This dataset includes reference point clouds (i.e., as the target cloud), including accurately georeferenced \textit{ALS} and \textit{MLS} data, and an \textit{MLS} point cloud that needs to be registered (i.e., as the source cloud). In this paper, we focus on one of the main streets in Munich's central area, Sonnenstraße. The source cloud was collected using a Viametris MS-96 mobile laser scanner mounted on a bicycle \citep{ViametrisMS96}.

The target point clouds were acquired using \textit{MLS} and \textit{ALS}. \textit{ALS} data was acquired from 1600 m altitude, and a point density of over 25 points per square meter was achieved, with a maximum of 40 points in certain areas. The scanner was set at a 60° scan angle and covered a width of 1212m, with more than 60\% overlap. The \textit{MLS} data was acquired using a Z+F Profiler 9012 mobile scanner.

\subsection{Evaluation metrics of registration accuracy}
\label{sec:evaluation}

Since ground-truth transformation parameters for this dataset are not available, the registration accuracy is assessed by direct visual inspection and calculating the spatial distances in stable areas between registered point clouds. Evaluating registration accuracy mainly involves the following three steps:

\begin{itemize}
    \item \textbf{Patch selection} – Stable planar areas of \(\,1\,\text{m}^{2}\) are manually selected from mutually orthogonal planes on building façades and road surfaces. The distribution of these patches has already been validated by \textit{SSC} in \textit{MLS} data fragmentation, and it could ensure that the registration errors are probed throughout the scene.
    \item \textbf{Axis definition} – The selected patches have different orientations. In order to unify the direction of the registration error represented by the patch-based distance, we redefined the meaning of each patch orientation: the street direction is \(X\)-axis, the side street direction is \(Y\)-axis, and the vertical direction is \(Z\)-axis. Hence, the patch's surface normal defines which direction its distance indicates in terms of registration error.
    \item \textbf{Patch distance calculation} – For each patch, the \textit{M3C2} distance between corresponding patches in the reference point cloud and the registered \textit{MLS} point cloud is computed along the patch normals. The \textit{M3C2} distance metric is adopted for its robustness against non-uniform point spacing and noise \citep{yang2023towards, yang2025piecewise}. For each axis, the \textit{M3C2} values of all patches are averaged to yield an axis-wise registration error. The mean error in all three axis directions is used as the overall registration accuracy.
\end{itemize}

\subsection{Results of point cloud fragmentation }
\label{Point Cloud Fragmentation }

\begin{figure}
	\centering
	\includegraphics[width=.8\columnwidth]{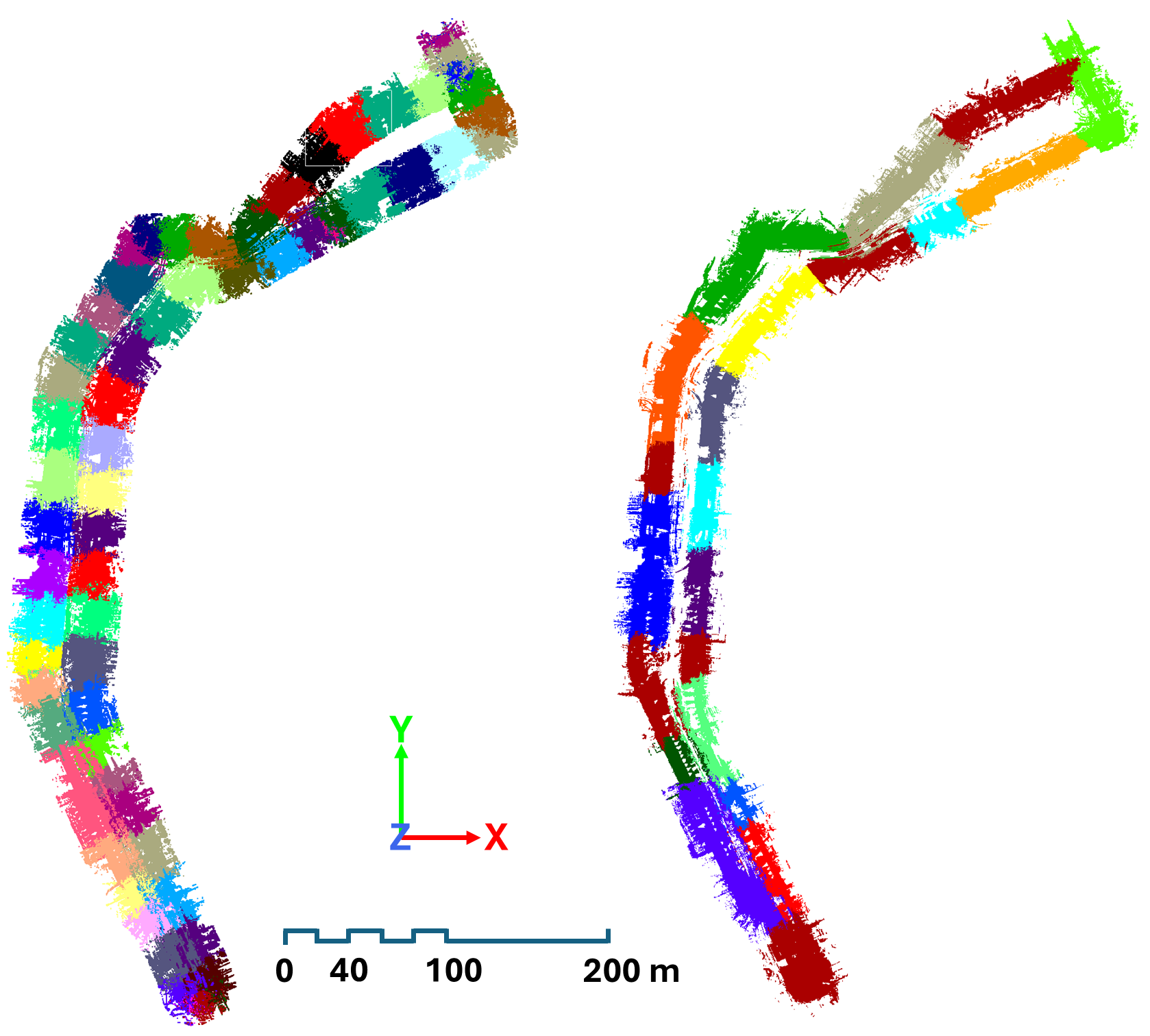}
	\caption{Results of \textit{MLS} data fragmentation: The left figure shows that the point cloud is initially divided into 10 s frames, and the right figure shows adaptive fragments after applying \textit{SSC}.}
	\label{Result_fragments}
\end{figure}

After pre-processing (Section~\ref{sec:data-prep}), dynamic objects (cars, vegetation, pedestrians) were removed from both the target and source point clouds. Resulting in a reduction of over 60\% of the total points. Retaining only static objects can not only enhance computational efficiency but also reduce the incorrect correspondences established in coarse registration.
The preprocessed \textit{MLS} point cloud spans approximately 2.5 kilometers and contains over 225 million points. After applying the Semi-sphere Check (\textit{SSC}) validation, the fragmentation results are presented in Figure \ref{Result_fragments} (right). The complete \textit{MLS} point cloud is finally divided into 22 adaptive fragments that contain sufficient geometric features while being restricted in length.

\begin{figure}
\centering
\begin{subfigure}[b]{0.5\textwidth}
    \includegraphics[width=\linewidth]{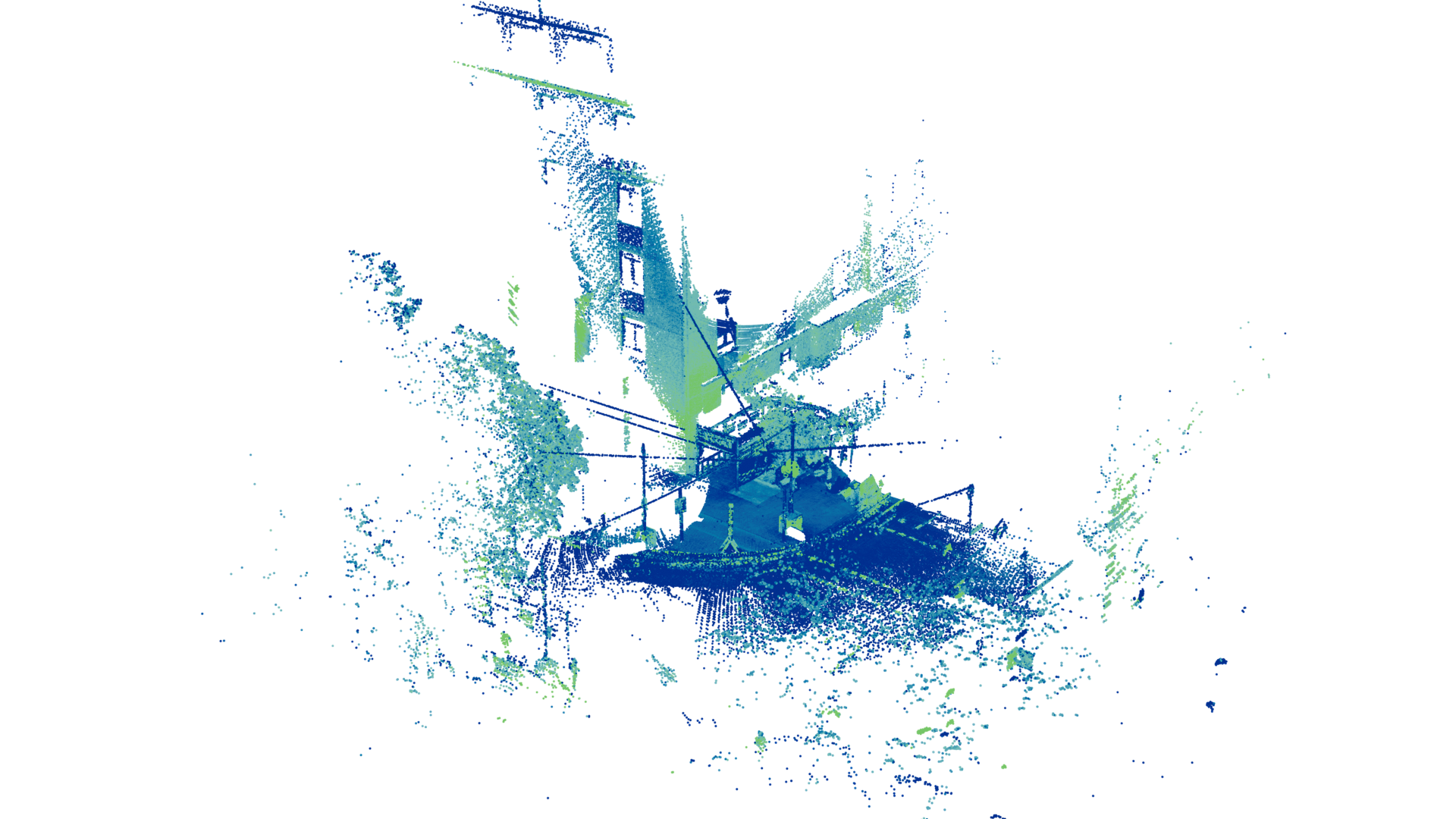}
    \caption{10 s Fragment}
    \label{exp_fragment_a}
\end{subfigure}\hfill
\begin{subfigure}[b]{0.5\textwidth}
    \includegraphics[width=\linewidth]{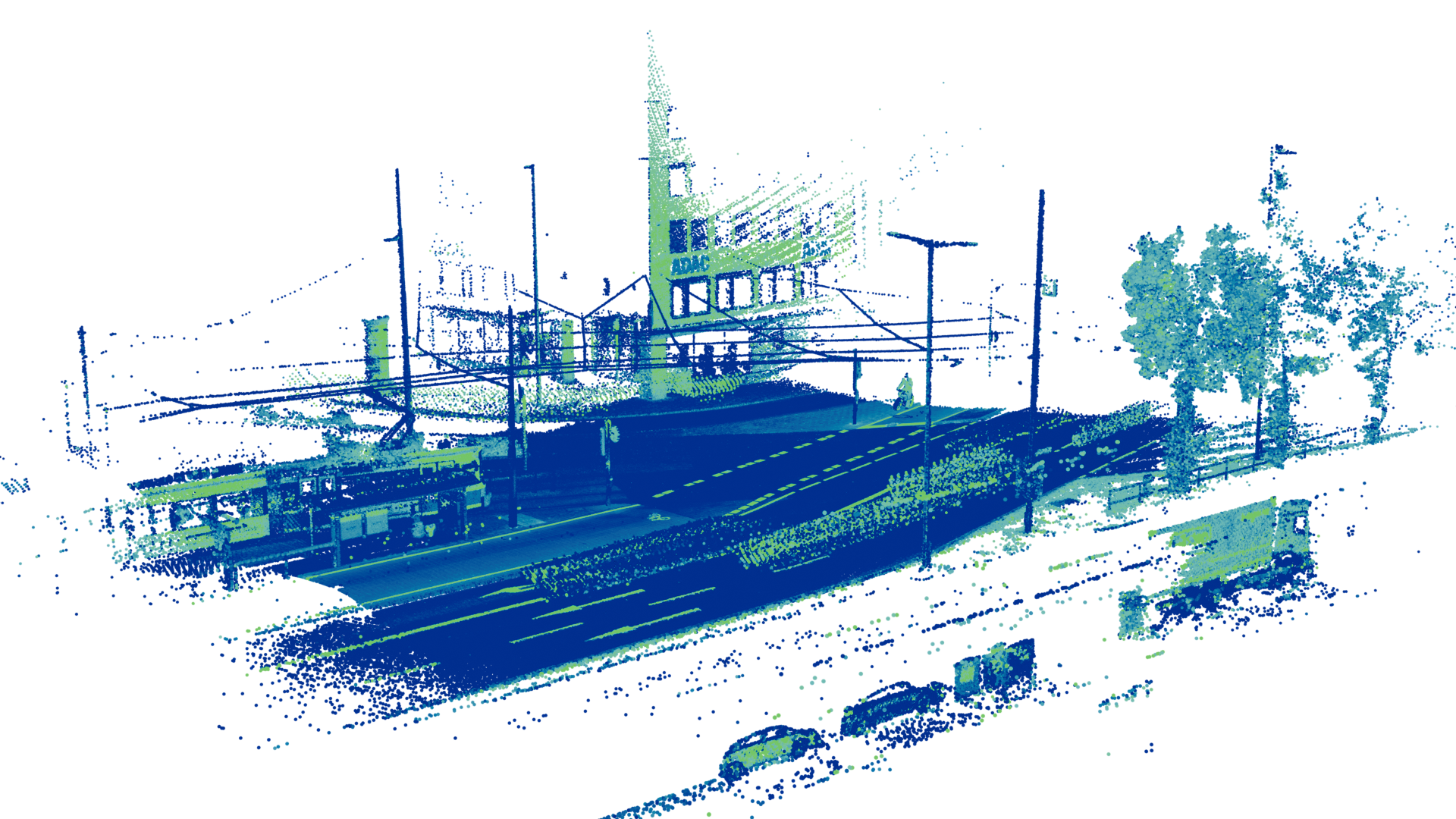}
    \caption{20 s Fragment}
    \label{exp_fragment_b}
\end{subfigure}
\caption{Examples of segments rejected by the \textit{SSC} due to insufficient planar geometry: (a) static capture scenario with limited spatial coverage; (b) segment dominated by horizontal surfaces, lacking sufficient orthogonal planar information.}
\label{example_fragments}
\end{figure}

It can be seen that the length of each adaptive segment is different. This is due to the lack of buildings (planar surfaces) in some initial fragments. Figure \ref{example_fragments} shows two fragments with fixed time intervals. In these cases, the \textit{SSC} process reports that these fragments do not contain enough mutually orthogonal surfaces. 


\subsection{Results of coarse registration}

\begin{figure}
    \centering
    \begin{tabular}{c c c}
        \toprule
        & \textbf{Unregistered fragments} & \textbf{Coarse registration by \textit{GROR}} \\
        \midrule
        & \includegraphics[width=0.20\textwidth]{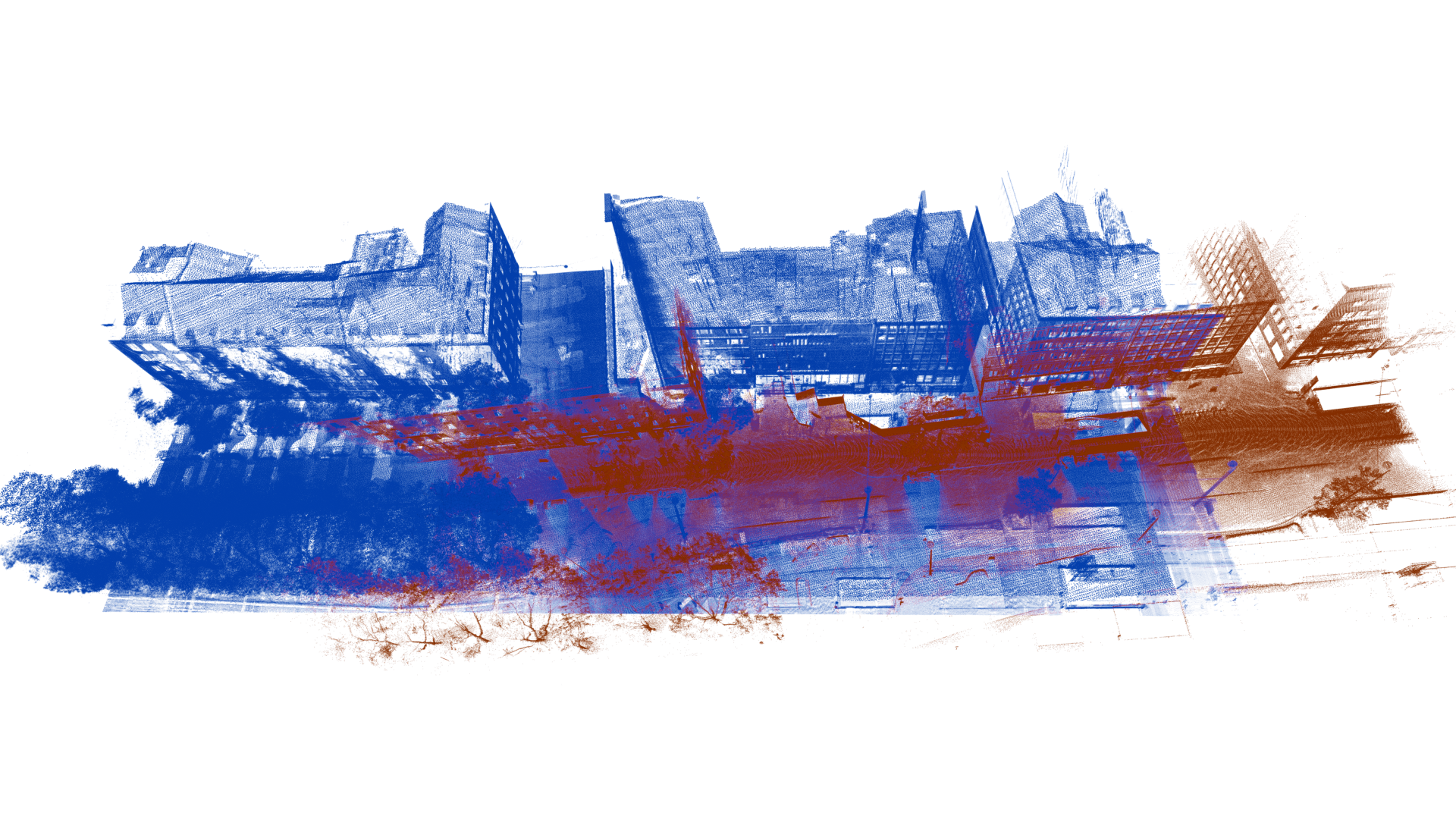}
        & \includegraphics[width=0.20\textwidth]{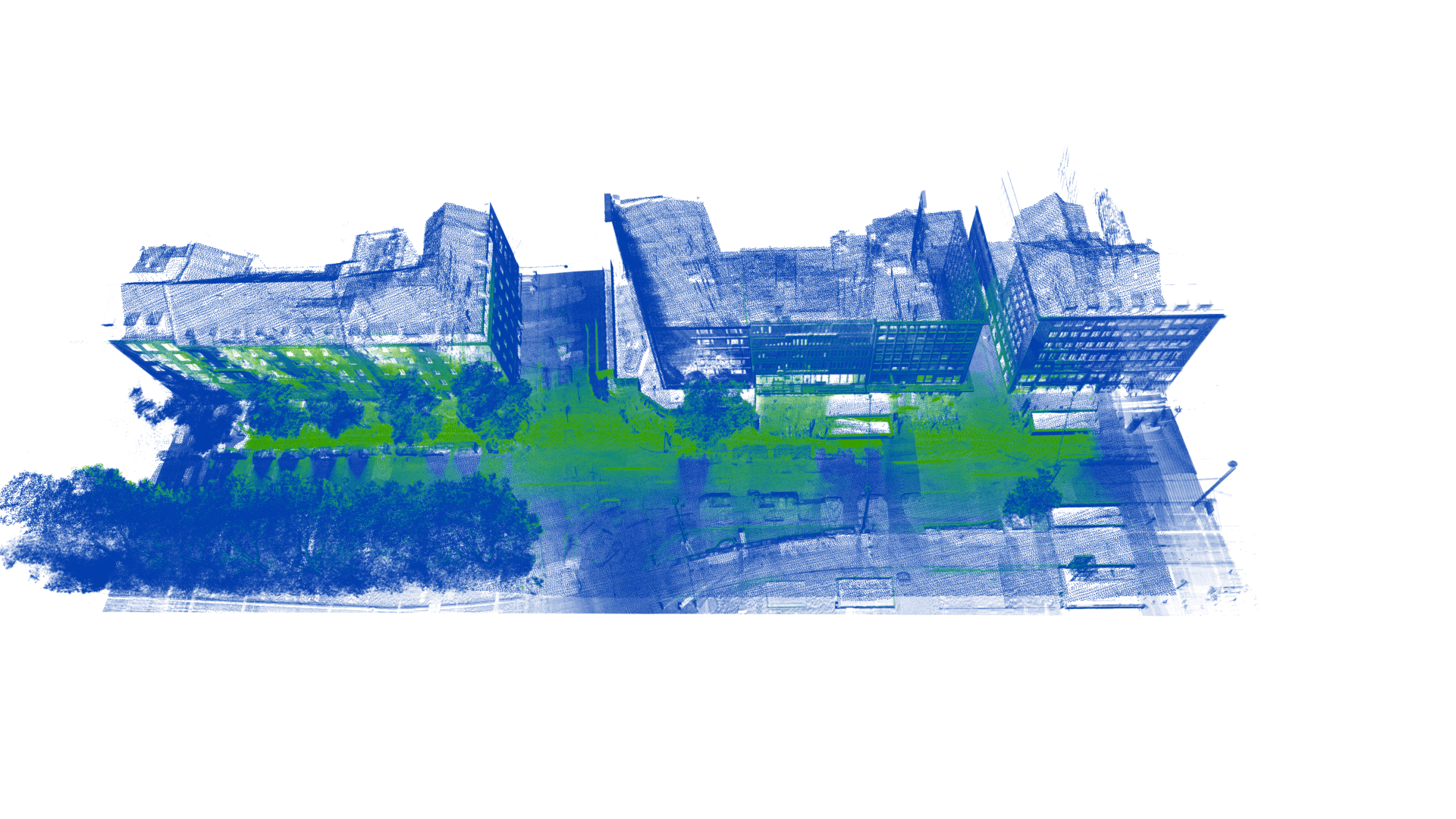   } \\

        & \includegraphics[width=0.20\textwidth]{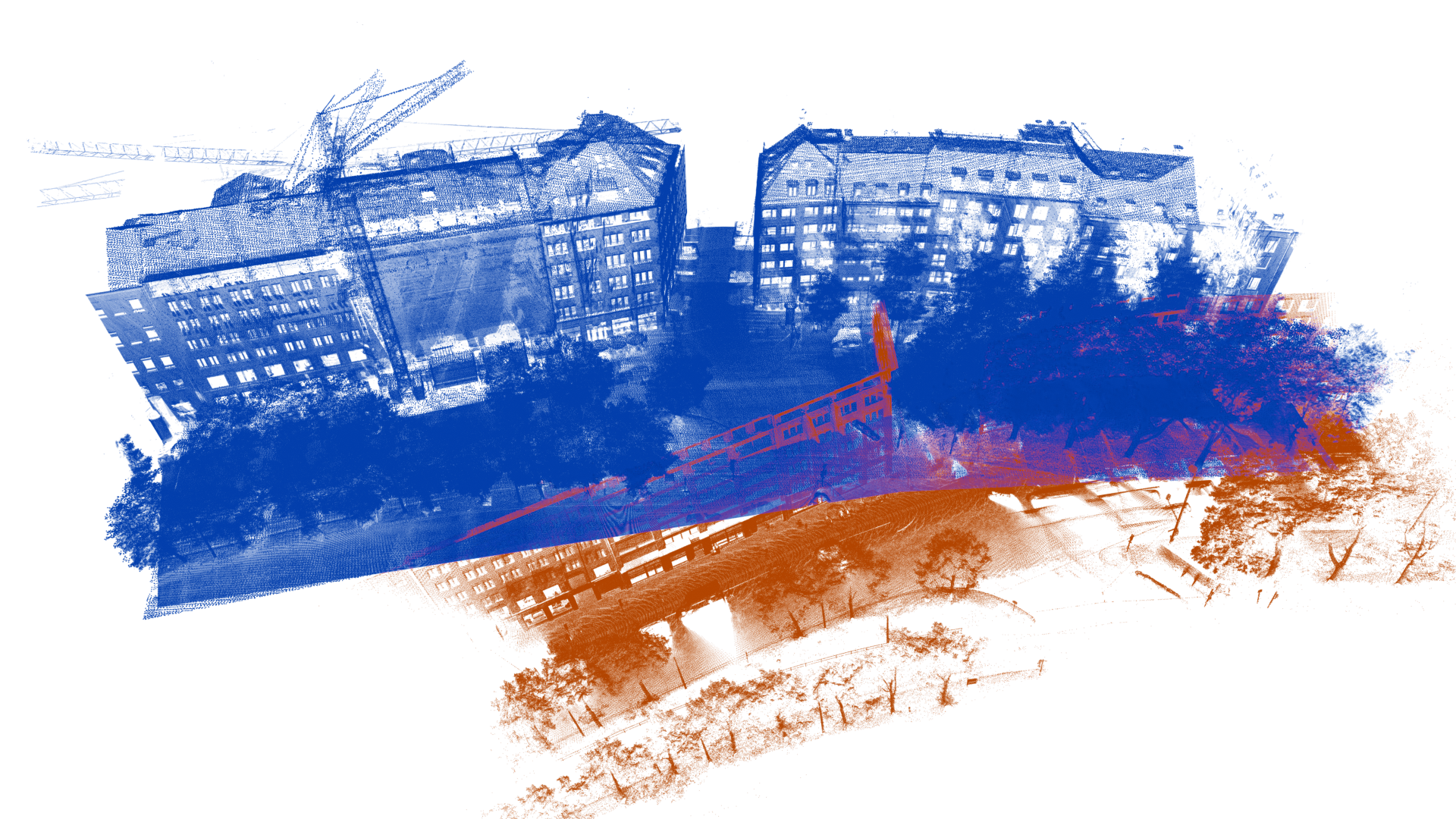}
        & \includegraphics[width=0.20\textwidth]{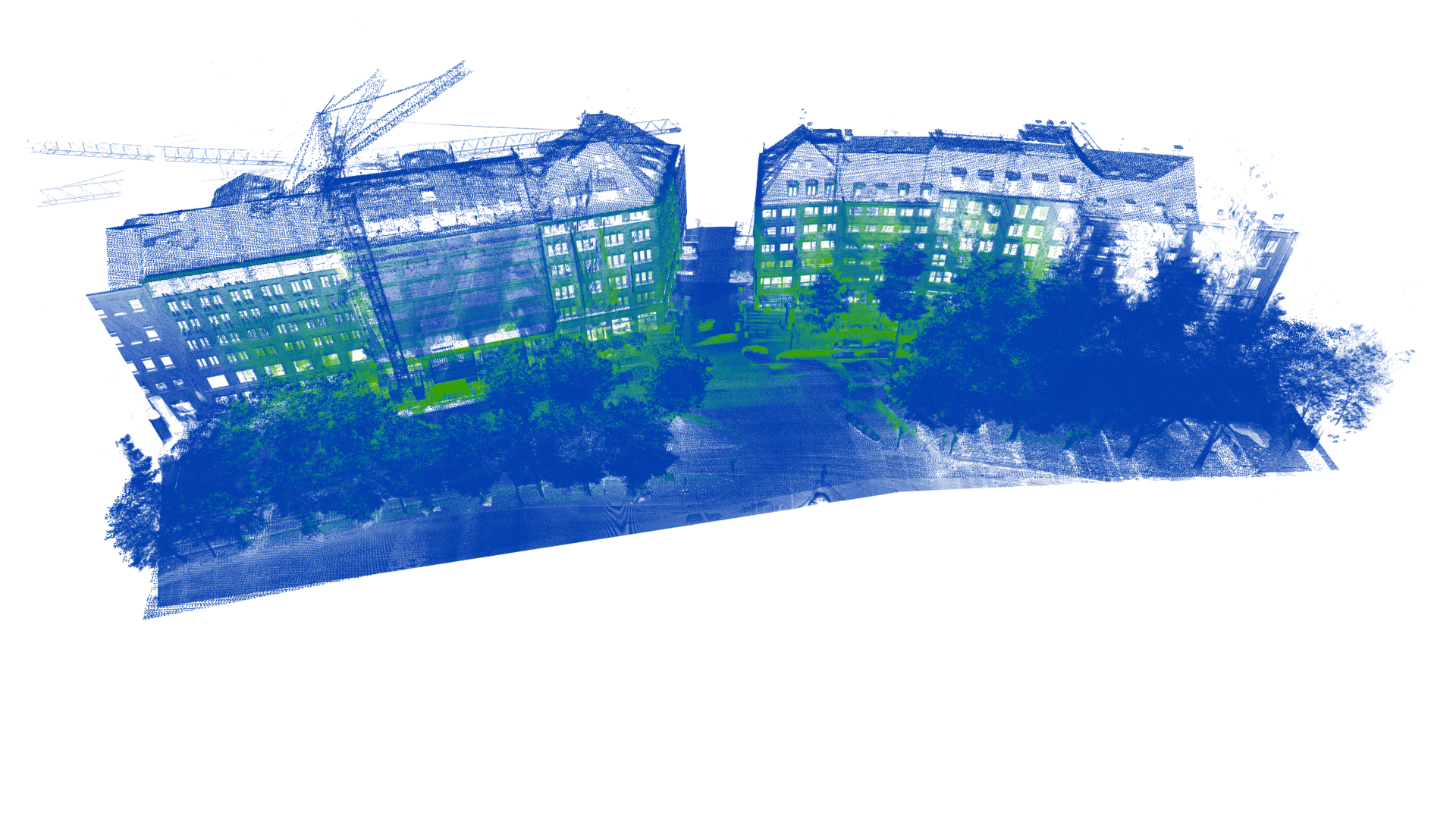} \\

        & \includegraphics[width=0.20\textwidth]{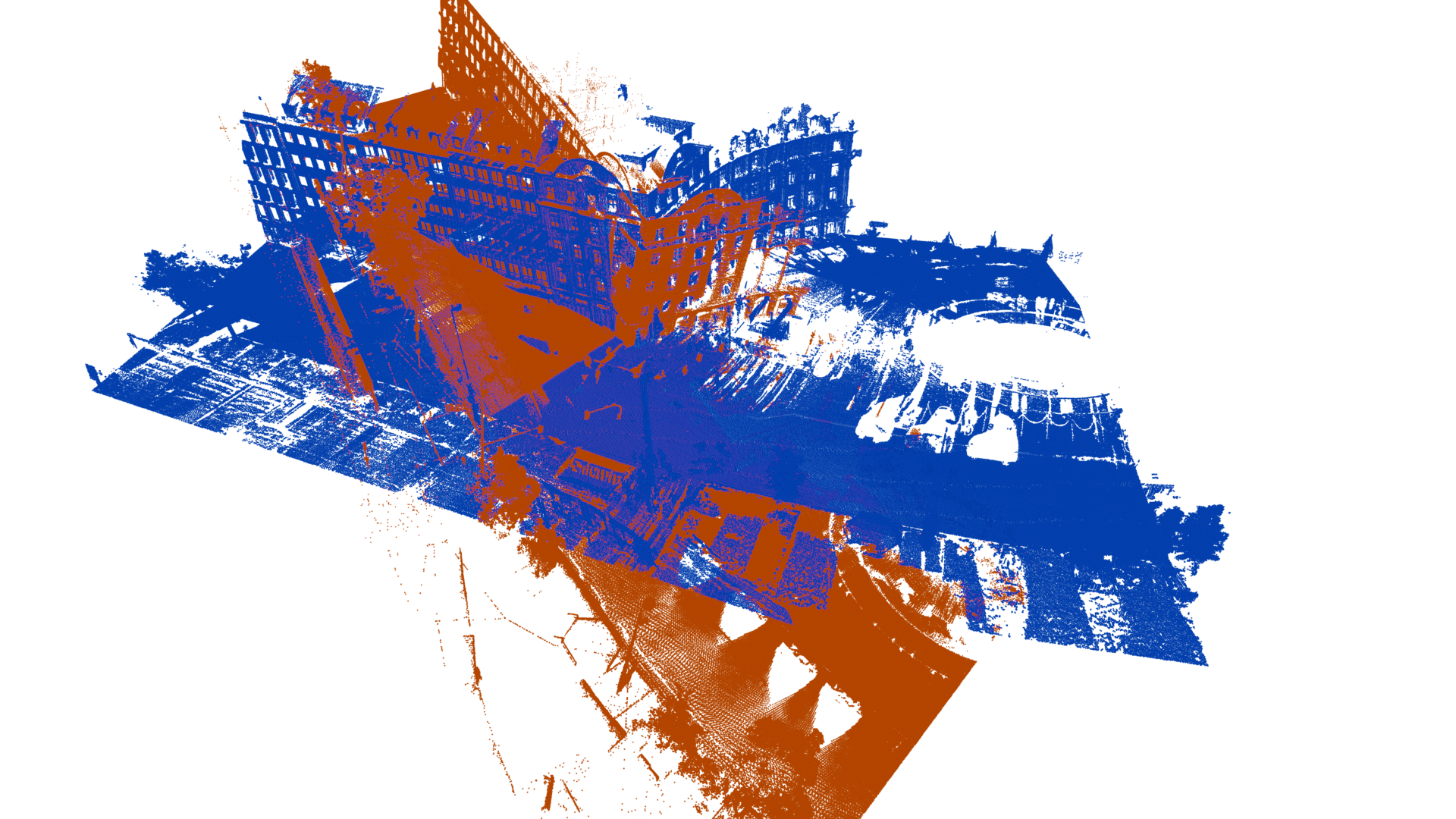}
        & \includegraphics[width=0.20\textwidth]{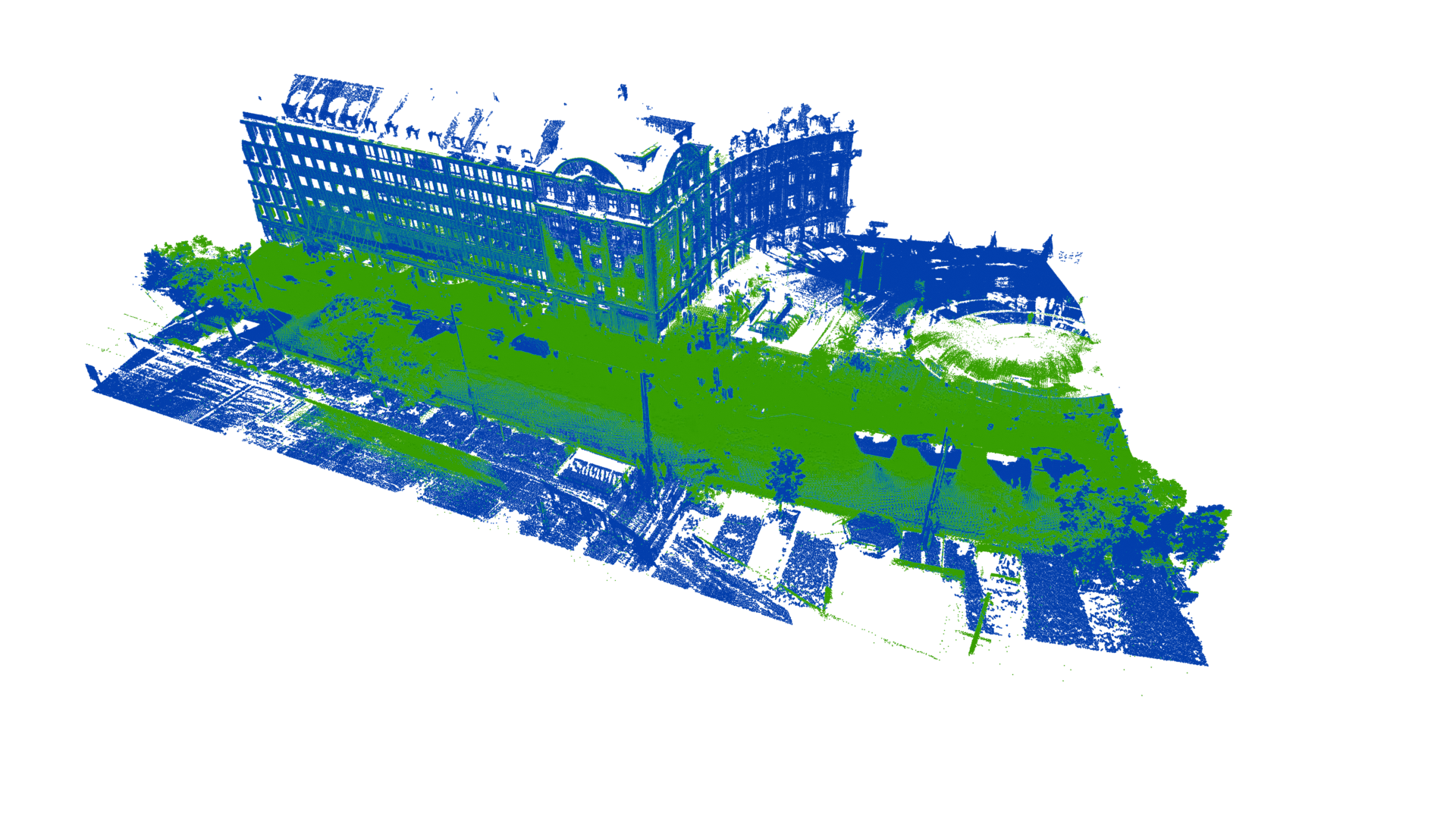} \\

        & \includegraphics[width=0.20\textwidth]{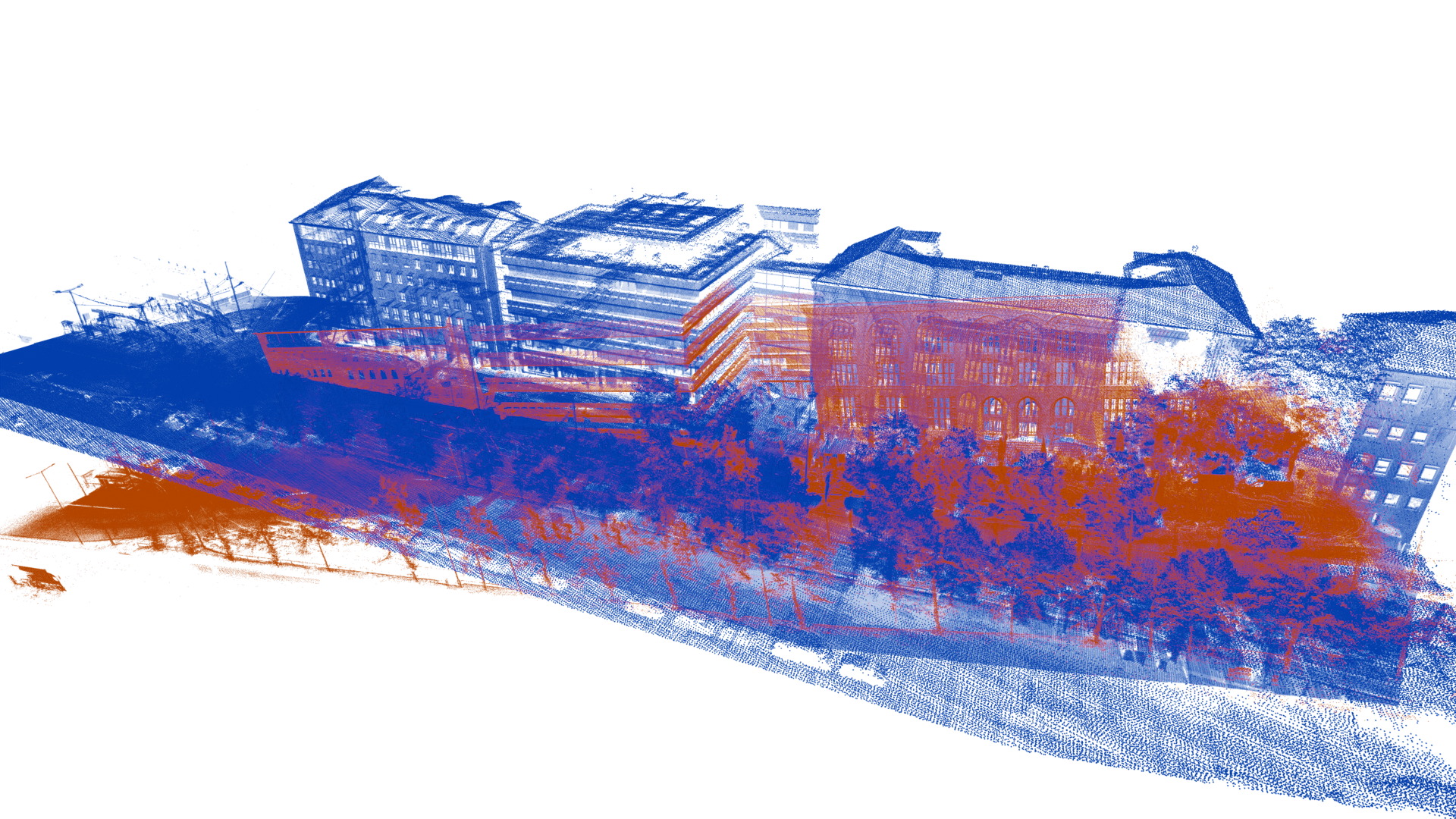}
        & \includegraphics[width=0.20\textwidth]{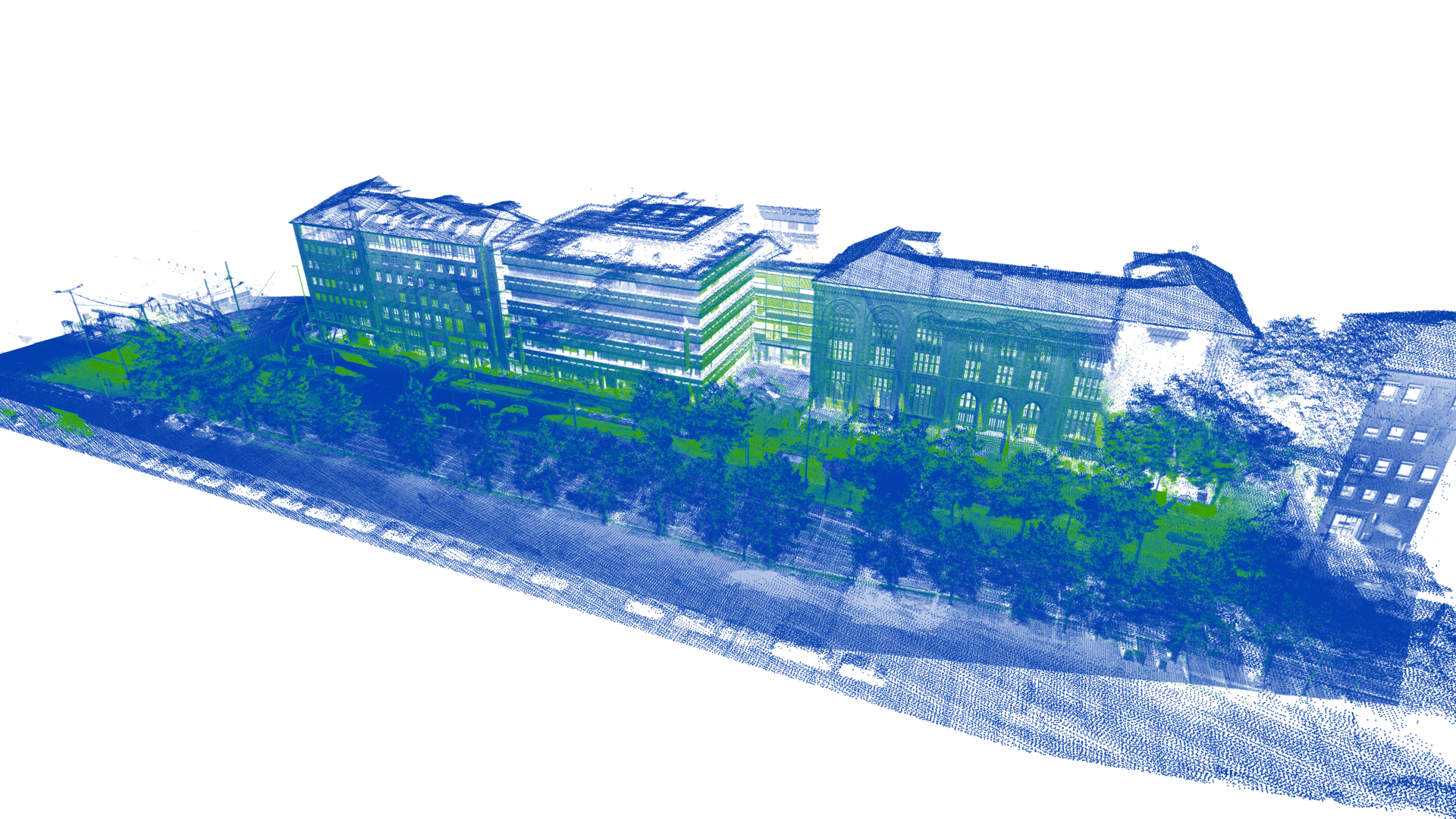} \\
        \bottomrule
    \end{tabular}
    \caption{Coarse registration results of four selected fragments validated by \textit{SSC}. Each scenario illustrates the alignment between the target point cloud (blue), the source point cloud (orange), and the registered cloud (green).}
    \label{fig:gror4}
\end{figure}

\textit{GROR} method is applied to the generated 22 fragments separately for initial alignment \citep{yan2022gror}. Figure \ref{fig:gror4} displays four fragments selected from the coarse registration results. From the visualized results, the source point clouds are all successfully aligned to the target point cloud, despite varying street scenarios. This coarse registration results in a reliable initial alignment of \textit{MLS} data to the reference point cloud, which allows subsequent fine registration to avoid local minima.

\subsection{Results of fine registration}
\label{Results of fine registration}

As mentioned in Section~\ref{sec:fine}, significant deviations may still exist between coarsely registered point clouds. In some fragments, these deviations in stable areas can achieve $0.1 m$. The proposed \textit{PV-GICP} is applied to refine the registration results using validated planar voxels.

\begin{figure}
\centering
\includegraphics[width=\linewidth]{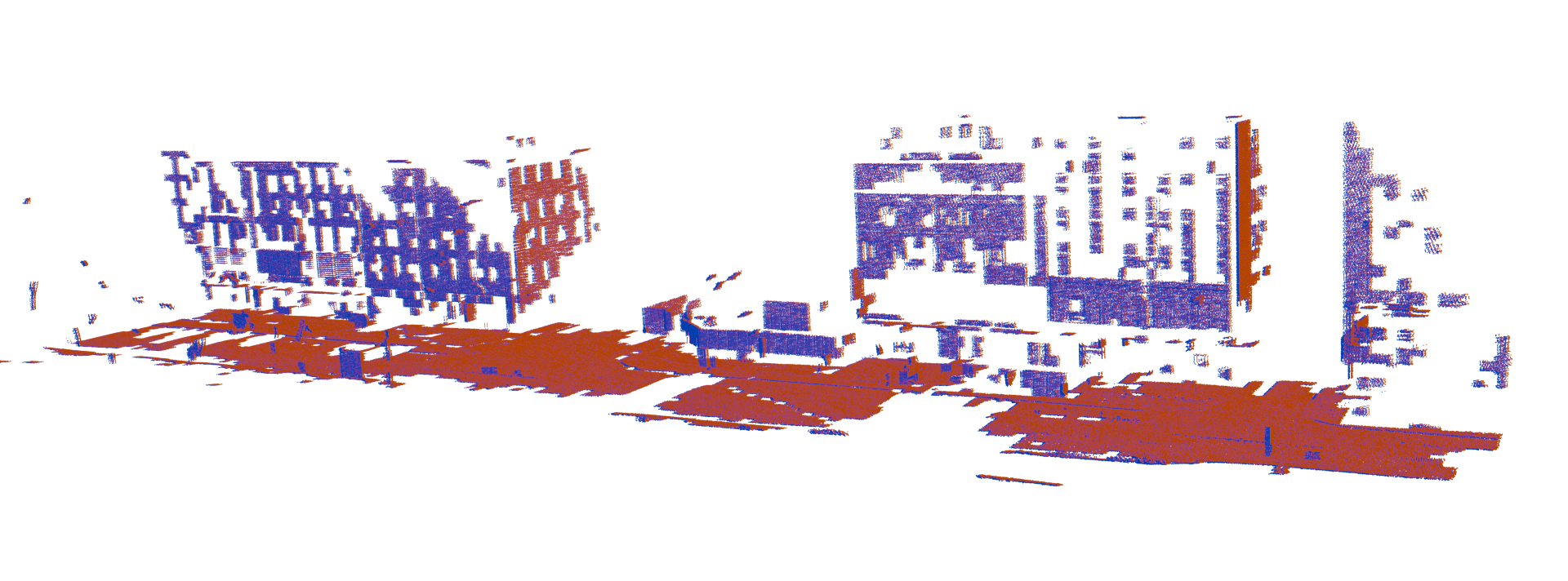}
\caption{Identified planar regions within selected voxels. Blue and red parts are the target and source clouds, respectively.}
\label{fig:planes}
\end{figure}

Figure \ref{fig:planes} presents a coarsely registered fragment, showing identified planar voxels through \textit{PV-GICP}. This figure illustrates another significant benefit of utilizing the \textit{SSC} for \textit{MLS} processing. It can be observed that these planar areas used for subsequent fine registration are mainly located on the building facades and road surfaces, which is in line with the assumption of stable regions.

\begin{figure*}
\centering
\includegraphics[width=1 \textwidth]{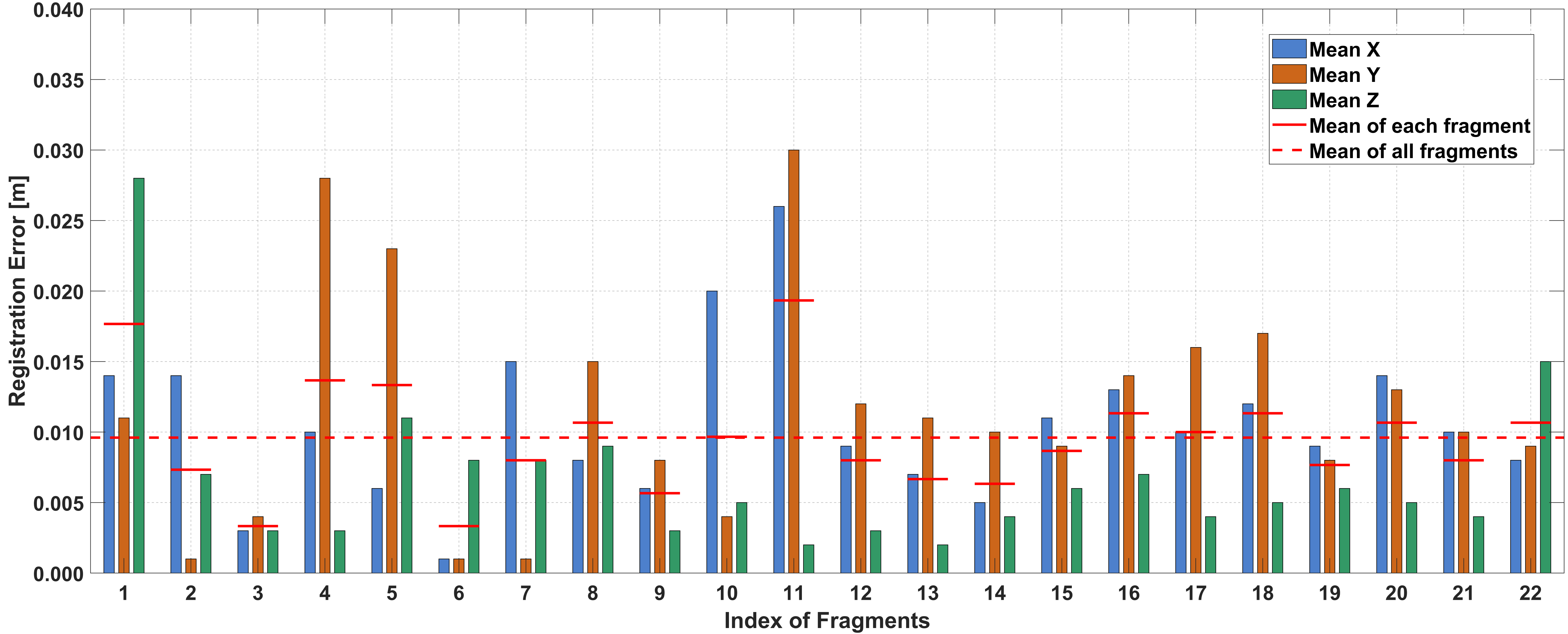}
\caption{Accuracy evaluation of fine registration across all 22 fragments using \textit{PV-GICP}. Each bar represents the mean \textit{M3C2} distance computed from selected patches extracted along respective defined axes. Solid red lines indicate the mean registration error per fragment, and the dashed red line represents the overall average registration error.}
\label{FIG:fineR}
\end{figure*}

After performing \textit{GICP} on identified planar voxels, we compute \textit{M3C2} distances on selected patches to quantify the registration errors. Since each fragment has been successfully validated by \textit{SSC}, these fragments should contain planar patches along all three orthogonal axes. Leveraging this advantage, 20 patches are extracted for each axis in each fragment. Figure \ref{FIG:fineR} shows the mean error of selected patches for each axis in each fragment. The average 3D registration error of each fragment and the overall mean error of the entire \textit{MLS} point cloud are also given. 

As can be seen from Figure \ref{FIG:fineR}, the registration accuracy of each fragment can achieve centimeter-level. Specifically, the errors along the $X$ and $Y$ axes are higher than those along the z-axis in most fragments. This is due to higher point density and quantity of the road surfaces than the building facade. In general, the average registration error of the entire \textit{MLS} point cloud is below 0.01 m, which can satisfy most applications of 3D urban model updates.

\begin{figure}
\centering
\includegraphics[width=\linewidth]{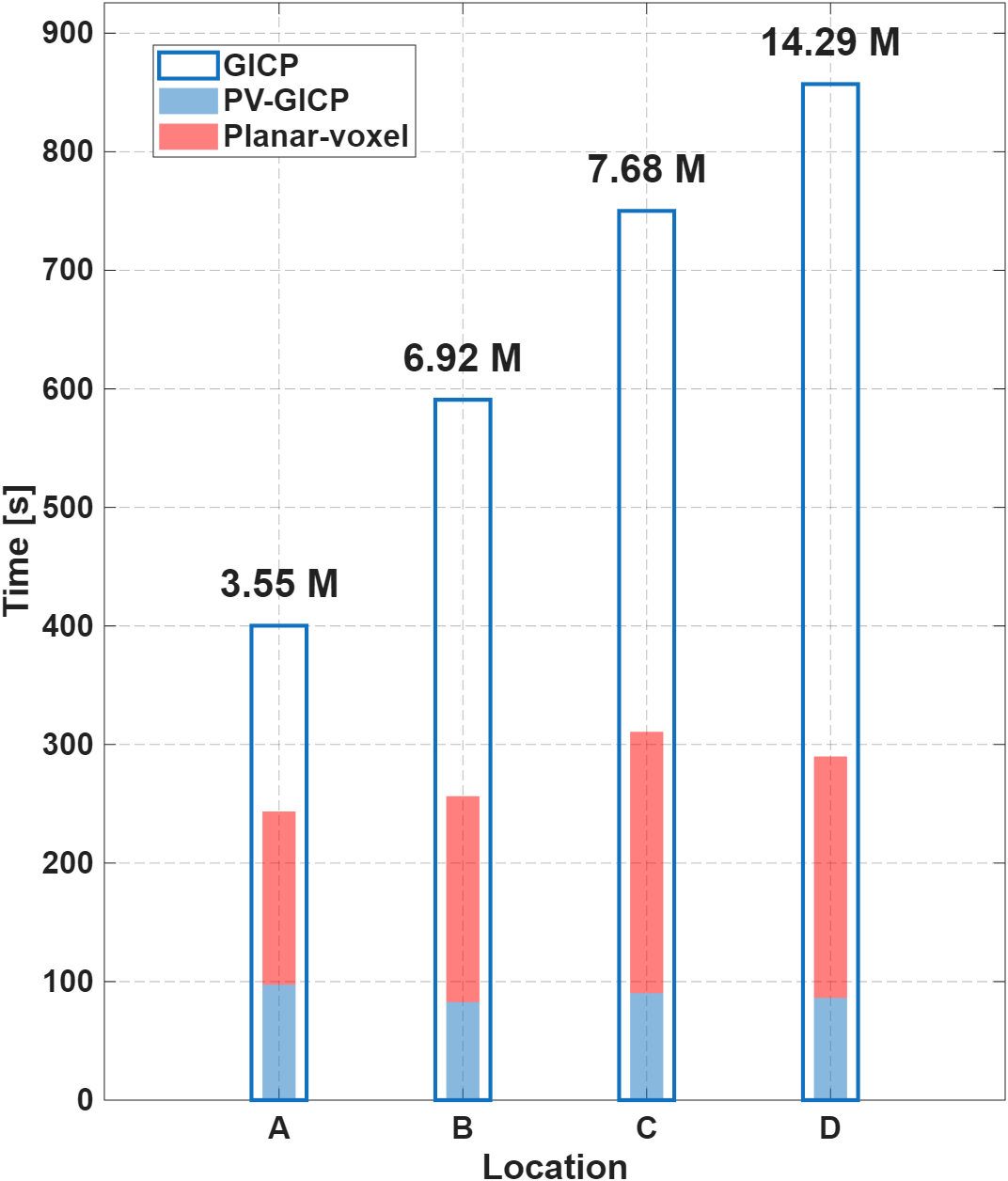}
\caption{Comparison of processing time between \textit{GICP} (unfilled blue bars), \textit{PV-GICP} approach (filled blue bars), and planar voxel extraction (filled red bars).}
\label{fig:time}
\end{figure}

Additionally, \textit{PV-GICP} can significantly decrease the computation time by removing non-planar regions. Figure \ref{fig:time} illustrates the processing times of the four fragments illustrated in Figure \ref{fig:gror4}. The unfilled blue bars represent the runtime when directly using standard \textit{GICP} for fine registration \citep{segal2009gicp}. In contrast, the filled blue bars indicate the runtime required for \textit{GICP} after extracting planar voxels. The red bar shows the computing time for the planar voxel extraction process. Additionally, the number displayed on top represents the point count in millions. It can be seen that the time for extracting planar voxels combined with the subsequent \textit{GICP} is less than half the runtime of using \textit{GICP} directly for fine registration in most cases. The difference in processing times varies depending on the size of the point cloud; larger clouds tend to yield greater time savings.

\subsection{Drift analysis} \label{Drift analysis}

The purpose of fragmenting the \textit{MLS} point cloud is to mitigate the impact of point cloud drift on registration accuracy caused by long trajectories that most times suffer from inaccuracies in the MLS trajectory estimation process that increases with time and trajectory length (\citep{XU2025PL4U}) . Conversely, the derived registration parameters of each fragment can reflect the extent of \textit{MLS} point cloud drift. This section presents the time-dependent drift detected via the proposed workflow and highlights the superiority of \textit{SSC}-based fragmentation over fixed-interval fragmentation.

\begin{figure*}
\centering
\includegraphics[width=1 \textwidth]{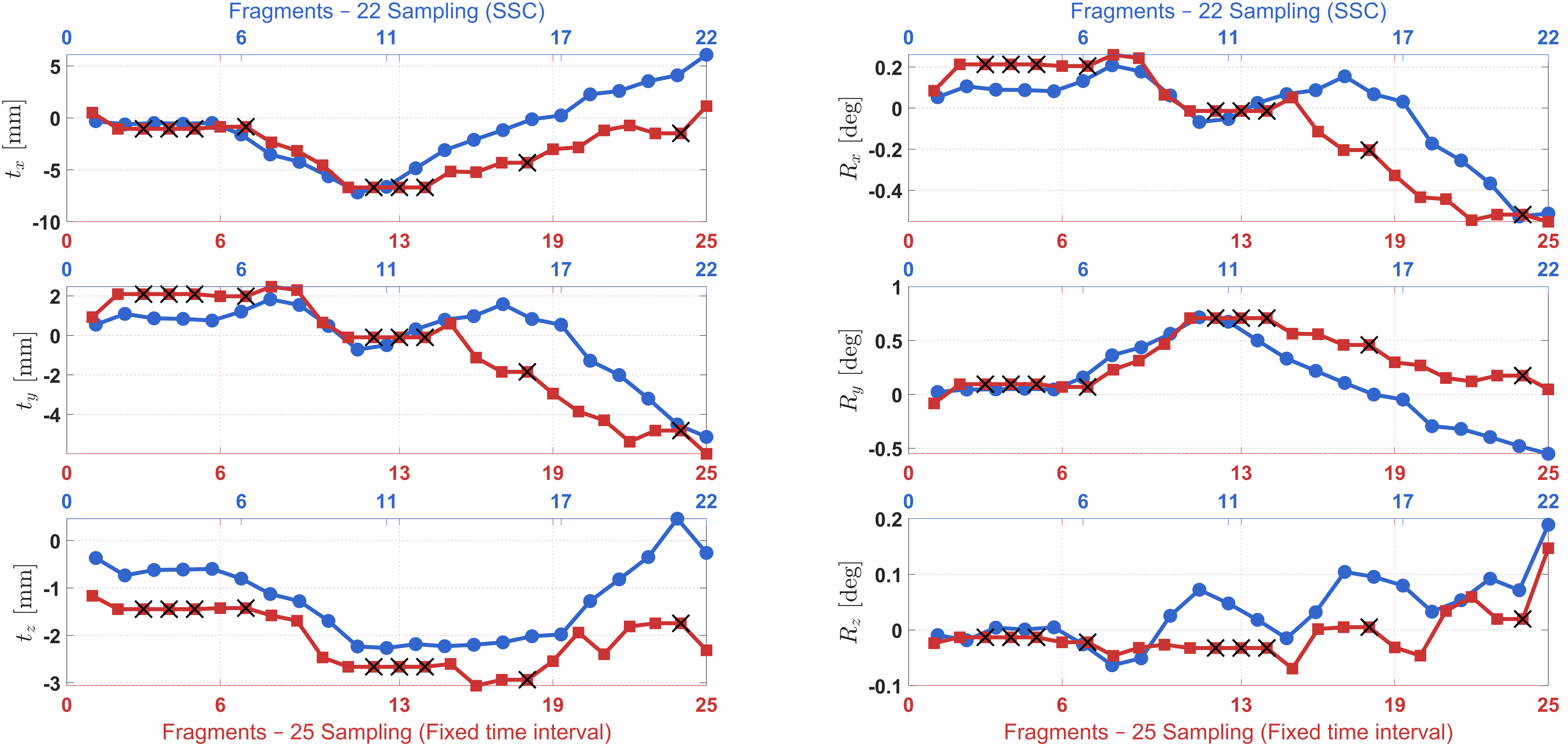}
\caption{Comparison of translation and rotation for each fragment obtained using the \textit{SSC} (blue) and fixed time interval segmentation (red). The left column displays cumulative translations along the three coordinate axes, while the right column shows cumulative rotation angles. The bottom and top axes represent the fragment indices corresponding to the \textit{SSC}-based and fixed time interval-based fragmentation methods, respectively. Black crosses denote fragments for which registration failed or yielded invalid results.}
\label{FIG:6plot}
\end{figure*}

By analyzing the transformation matrices obtained for all fragments that passed the \textit{SSC}, we extracted both rotation angles ($R_x$, $R_y$, and $R_z$) and translation components ($t_x$, $t_y$, and $t_z$), as illustrated in Figure~\ref{FIG:6plot}. These transformation parameters are relative to the first fragment starting from the \textit{MLS} origin, thus showing the accumulated drift effects. We also compare with the transformation parameters derived based on the fragments of fixed 30s intervals. Thus, there are 22 frames validated by the \textit{SSC} method (in blue) and 25 frames by a fixed 30 s interval (in red). Fixed-interval fragments that failed to be registered due to insufficient geometric features have their transformations interpolated based on the results of their adjacent fragments.

Figure~\ref{FIG:6plot} offers intuitive observations of the gradual drift over time. For instance, by the tenth frame, a significant translation trend in the x-direction, which decreases by approximately 7mm, can be clearly seen. After the tenth frame, $t_x$ increases and becomes positive. Similarly, the rotation angle along the y-axis exhibits a trend of first increasing and then decreasing. This trend of drifting and then recovering is mainly caused by the loop closure detection in the SLAM (simultaneous localization and mapping) algorithm used in \textit{MLS} point cloud generation \citep{Slam1,Slam2}. \\
An examination of the fragments without using \textit{SSC} reveals that the translations display a more erratic behavior. In the accompanying plots, black crosses indicate the fragments where registration fails, typically attributable to a lack of sufficient features. Besides, an increase in fluctuations can be observed in translations in the y- and z-directions.

\begin{figure*}
	\centering
	\includegraphics[width=1 \textwidth]{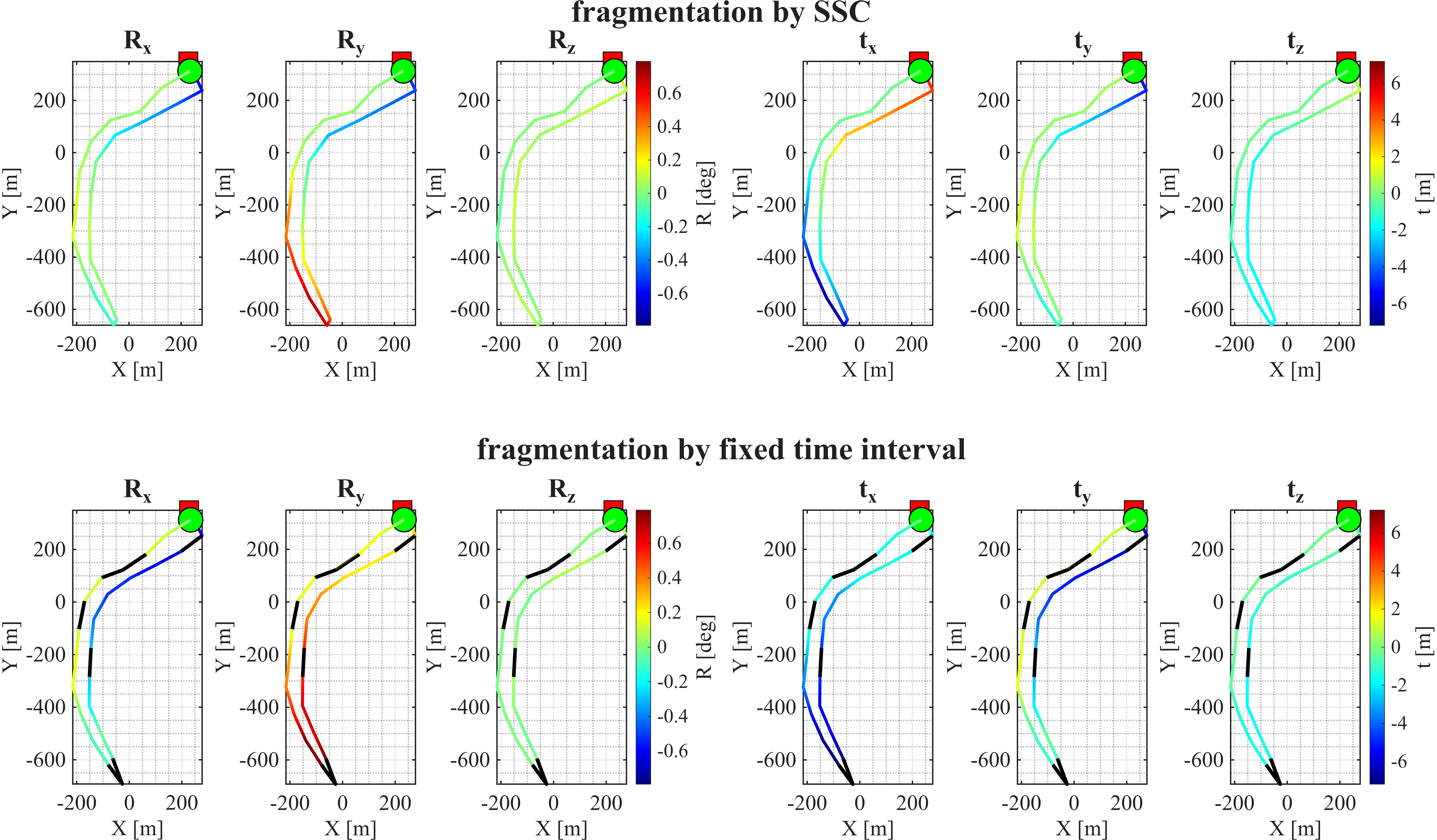}
	\caption{Calculated drifts derived from consecutive transformations of each fragment. Each panel shows a color-coded drift representation on the \textit{MLS} trajectory, highlighting three rotation angles ($R_x$, $R_y$, $R_z$) and three translation components ($t_x$, $t_y$, $t_z$) across two fragmentation strategies. The first row represents the \textit{SSC}-based results, and the second row represents the results using a fixed time interval.}
	\label{FIG:trajec}
\end{figure*}

Figure~\ref{FIG:trajec} illustrates the trajectories imposed by cumulative rotations and translations in Figure~\ref{FIG:6plot}. This figure generally presents the spatial distribution of derived drift effects for each fragment generated by the \textit{SSC} method and a fixed time interval.

As depicted in Figure~\ref{FIG:6plot}, the plots demonstrate a maximum deviation around the tenth fragment, which corresponds to the higher drifts at the U-turn area for the \textit{MLS} trajectory in Figure~\ref{FIG:trajec}. Initially, these transformation parameters increase over time (especially for $R_y$ and $t_x$), but following the U-turn, a decrease occurs owing to the loop closure corrections. After the 17th fragment, however, loop closure becomes unfeasible due to environmental obstacles, leading to a renewed increase in drift.

The drift results of fragments based on fixed time intervals are shown in the second row of Figure~\ref{FIG:trajec}. The areas of invalid registrations are highlighted with black sections. It can be seen that a significant proportion of fragments fail to be registered or exhibit large drift values.

A detailed examination of the trajectory could reveal deeper insights. In Figure~\ref{FIG:trajec}, the starting point is marked by a green circle and the endpoint by a red square. Initially, the errors are relatively low; however, after approximately one kilometer, there is a significant increase in drift. After the U-turn area, these drifts are obviously reduced. This stage demonstrates the crucial role that loop closure detection plays in mitigating drift effects in the \textit{MLS} system. In the final section, due to environmental obstructions, there was little overlap between the scanned area and the previously scanned area, resulting in a gradually increasing drift.

\begin{figure}
\centering
\includegraphics[width=1\columnwidth]{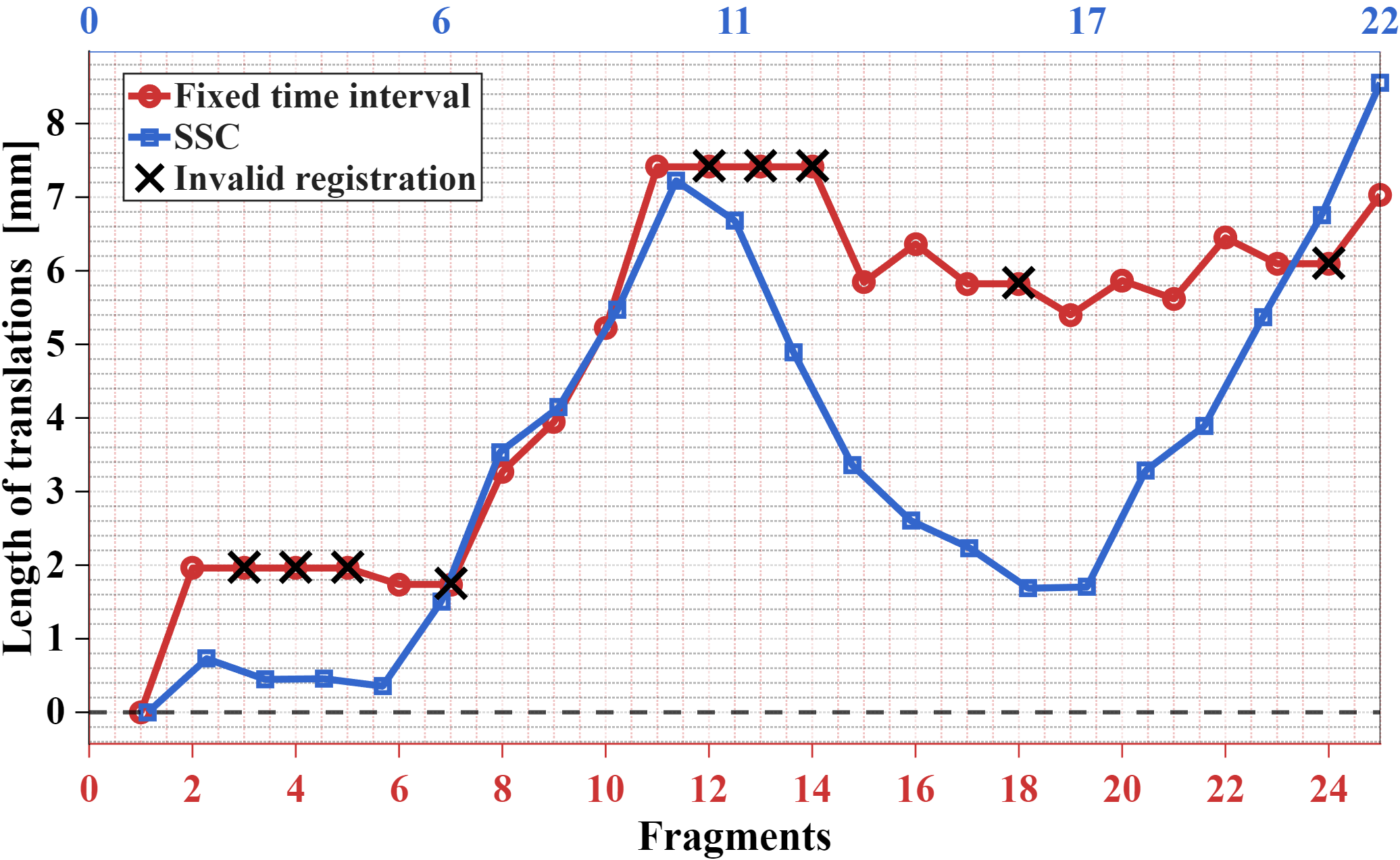}
\caption{Comparison of the derived translation length across consecutive fragments between the \textit{SSC}-based fragmentation (blue) and a fixed-time interval segmentation (orange). Black crosses mark fragments with invalid registrations due to insufficient geometric features.}
\label{FIG:tra}
\end{figure}

As a summary of the advantages of using \textit{SSC}-based fragmentation, the cumulative 3D translation magnitudes are plotted in Figure \ref{FIG:tra}. It can be seen that an increasing trend occurs by the tenth fragment, followed by a decreasing trend until the 17th fragment. Afterward, an increasing drift appears again until the trajectory ends. These varying trends of translations agree with the assumption of the drift behaviors in the \textit{MLS} point cloud. In contrast, the fragments using fixed-time segmentation do not show this behavior after the tenth fragment.

\section{Discussion}
\label{Discussion}
In this section, we discuss the practical implementation of our \textit{SSC} adaptive fragmentation and \textit{PV-GICP} pipeline. Section \ref{adaptive} outlines the advantages of adaptive fragmentation. Section \ref{parameter} discusses parameter sensitivities and recommended settings (e.g., angular threshold in \textit{SSC} and voxel size in \textit{PV-GICP}. Section \ref{limitations} delineates the method’s limitations and scope, especially in scenes lacking dominant planar façades or with predominantly curved geometry.

\subsection{Benefits of adaptive fragmentation}
\label{adaptive}
The proposed \textit{SSC}-based fragmentation yields clear advantages not only over conventional fixed time interval segmentation \citep{XU2025PL4U} but also trajectory partitioning based on the Douglas–Peucker algorithm \citep{lucks2021improving} for \textit{MLS} point cloud registration. Since each fragment generated by \textit{SSC} contains sufficient mutually orthogonal surfaces, these geometries can generate more edges and corner points, resulting in successful coarse registration by \textit{GROR}. Well chosen initial alignments further prevent local minima in the following fine registration. Therefore, high-quality fragments with adaptive size can facilitate robust and accurate registration of \textit{MLS} point clouds.

In addition to improving coarse alignment, the SSC-based fragmentation also enhances the performance of fine registration using PV-GICP. Our PV-GICP module also benefits directly from this fragmentation. Previous studies such as \citep{segal2009gicp}, \citep{yang2015goicp}, and \citep{yang2025piecewise} have introduced probabilistic and robust-kernel registration techniques that improve convergence but often increase computation time. In contrast, our preprocessing model reduces both runtime and the likelihood of registration failure by operating only on geometrically validated planar fragments.

Furthermore, it better demonstrates the drift effect of \textit{MLS} point clouds.  
Based on the derived transformation parameters of each adaptive fragment, the drift behaviors are intuitively presented on the \textit{MLS} trajectory, which corresponds to the practical scenarios. In contrast, the derived drifts by fragments with a fixed time interval generate noisy and incomplete results. Unlike previous geometric segmentation methods, SSC leverages the spatial orientation of planar surfaces to control fragment size adaptively. This orientation-aware fragmentation represents a novel strategy for mitigating MLS drift while maintaining registration robustness.

\subsection{Parameter setting}
\label{parameter}

Using fixed thresholds across all scenarios is suboptimal. In the used street \textit{MLS} datasets, point density varies significantly along the trajectory due to changes in vehicle speed and motion (e.g., stops, accelerations, and turns), even after applying voxel grid-based downsampling.

For \textit{SSC} validation, under-populated seeds should be discarded before proceeding. A minimum seed population is required to ensure reliability. The seed-direction dispersion is then computed using only the remaining seeds to determine whether they should be retained or rejected. These checks must be density-aware: in sparse or high-speed segments, the threshold of per-seed minimum can be relaxed (or fragments slightly extended) to avoid discarding valid façade directions. Conversely, in dense or slow segments, the minimum requirement should be increased and the dispersion threshold tightened to prevent accepting small and rough structures.

In fine registration, voxel size plays a critical role in \textit{PV-GICP} performance. When the voxel size is too small (e.g., 0.01 m), the data is partitioned into numerous tiny cells. In this case, planarity estimation becomes unstable, spurious micro-planes appear, and computation time increases sharply. If the voxel size is too large, multiple planes may be incorporated in one voxel, leading to biased normal estimates and thus removing useful areas. In urban environments with building façades, a voxel size of around 1 m is recommended when significant noise is present.

Similarly, the angular threshold in \textit{SSC} affects performance: tightening it (e.g., to 10°) increases accuracy in highly planar city blocks, while loosening it (e.g., to 20°) preserves more data in sparse or noisy regions, reducing failure rates at the expense of a few millimeters registration errors. Empirically, across all scenarios, a consistency ratio of 70\% proved robust.

\subsection{Applicability and limitations}  
\label{limitations}
In this article, we address registration, validation, and drift correction. As a representative example, we present results from Sonnenstraße, one of the important inner-city streets in Munich. In addition, we evaluated the workflow on multiple datasets, including dense urban street areas, bridges (Maximiliansbrücke), iconic traffic circles (Königsplatz), and even mountainous regions such as Höllentalklamm in Germany. Across these scenarios, the method consistently produced stable registrations. These tests indicate that the workflow is robust across varied street morphologies.

The relevant parameter settings in \textit{SSC} are currently still fixed empirical values, which may not be applicable to other street scenarios. To address this, more adaptable thresholds in \textit{SSC} can be established. Notably, the \textit{SSC}-based fragmentation strategy is particularly effective in urban environments. For scenarios lacking building facades, such as rural areas, the algorithm may generate longer fragments where internal drifts may still exist. 

The \textit{PV-GICP} algorithm proposed in this paper is based on the assumption that there are a large number of planar structures (such as building facades and road surfaces) in street scenes. Nevertheless, this limitation opens future research opportunities toward hybrid fragmentation strategies that integrate curved-surface recognition to complement the planar-based PV-GICP framework.

\section{Conclusions}
\label{Conclusion}

This article proposes a complete registration workflow for urban \textit{MLS} data, which introduces two novel components: an adaptive fragmentation strategy using Semi-sphere Check \textit{(SSC)} to generate locally rigid fragments with sufficient geometric features, and planar voxel-based \textit{GICP} \textit{(PV-GICP)} as an efficient fine registration that operates only on automatically identified planar areas.

Experiments on a 2.5 km \textit{MLS} dataset capturing Munich’s inner city areas demonstrate that the proposed workflow can achieve the registration accuracy better than 0.03 m, with 90\% of errors below 0.02 m, and it halves the computing time of fine registration compared to the state-of-the-art Generalized-\textit{ICP}. In addition, the derived transformation parameters can effectively reflect the drift behaviors of \textit{MLS} in an interpretable manner, thus providing an approach to assessing the data quality of the \textit{MLS} system.

In summary, the proposed \textit{MLS} point cloud registration pipeline offers a practical balance of accuracy, robustness, and computational efficiency, advancing the applications of targetless registration for 3D urban model construction and updates. Future work involves broadening the method to support multi-modal point clouds, incorporating color and intensity descriptors to enhance the registration of geometry-poor but texture-rich scenarios.

\section*{CRediT authorship contribution statement}
\textbf{Marco Antonio Ortiz Rincón}: Conceptualization,\\Methodology, Software, Investigation, Experiments, Formal analysis, Validation, Visualization, Writing – original draft, Writing – review \& editing. \textbf{Yihui Yang}: Conceptualization, Methodology, Formal analysis, Validation, Writing - original draft, Writing – review \& editing. \textbf{Christoph Holst}: Supervision, Writing - Review \& Editing, Funding acquisition, Project administration.

\section*{Declaration of competing interest}
The authors declare that they have no known competing financial interests or personal relationships that could have appeared to influence the work reported in this paper.

\section*{Acknowledgments}
We are thankful to Philipp-Roman Hirt and Florian Meßner (City of Munich / Landeshauptstadt München, Kommunalreferat – GeodatenService) for providing the point cloud datasets of the Sonnenstraße area in Munich.

\section*{Data and code availability}
To ensure reproducibility and long-term access, the repository contains the implementation of the SSC fragmentation, the PV-GICP, and example datasets demonstrating the workflow presented in this paper. Are openly available on GitHub at:
https://github.com/Marco-23/SSC-Registration.

\bibliographystyle{cas-model2-names}

\bibliography{cas-refs}

\begin{thebibliography}{42}
\expandafter\ifx\csname natexlab\endcsname\relax\def\natexlab#1{#1}\fi
\providecommand{\url}[1]{\texttt{#1}}
\providecommand{\href}[2]{#2}
\providecommand{\path}[1]{#1}
\providecommand{\DOIprefix}{doi:}
\providecommand{\ArXivprefix}{arXiv:}
\providecommand{\URLprefix}{URL: }
\providecommand{\Pubmedprefix}{pmid:}
\providecommand{\doi}[1]{\href{http://dx.doi.org/#1}{\path{#1}}}
\providecommand{\Pubmed}[1]{\href{pmid:#1}{\path{#1}}}
\providecommand{\bibinfo}[2]{#2}
\ifx\xfnm\relax \def\xfnm[#1]{\unskip,\space#1}\fi
\bibitem[{Aiger et~al.(2008)Aiger, Mitra and Cohen-Or}]{aiger2008fourpcs}
\bibinfo{author}{Aiger, D.}, \bibinfo{author}{Mitra, N.},
  \bibinfo{author}{Cohen-Or, D.}, \bibinfo{year}{2008}.
\newblock \bibinfo{title}{4-points congruent sets for robust pairwise surface
  registration}.
\newblock \bibinfo{journal}{35th International Conference on Computer Graphics
  and Interactive Techniques (SIGGRAPH'08)} \bibinfo{volume}{27}.
\newblock \DOIprefix\doi{10.1145/1399504.1360684}.
\bibitem[{Aoki et~al.(2019)Aoki, Goforth, Srivatsan and
  Lucey}]{aoki2019pointnetlk}
\bibinfo{author}{Aoki, Y.}, \bibinfo{author}{Goforth, H.},
  \bibinfo{author}{Srivatsan, R.A.}, \bibinfo{author}{Lucey, S.},
  \bibinfo{year}{2019}.
\newblock \bibinfo{title}{Pointnetlk: Robust \& efficient point cloud
  registration using pointnet}, in: \bibinfo{booktitle}{Proceedings of the
  IEEE/CVF Conference on Computer Vision and Pattern Recognition (CVPR)}, pp.
  \bibinfo{pages}{7163--7172}.
\bibitem[{Arroyo~Ohori et~al.(2018)Arroyo~Ohori, Biljecki, Kumar, Ledoux and
  Stoter}]{ArroyoOhori2018}
\bibinfo{author}{Arroyo~Ohori, K.}, \bibinfo{author}{Biljecki, F.},
  \bibinfo{author}{Kumar, K.}, \bibinfo{author}{Ledoux, H.},
  \bibinfo{author}{Stoter, J.}, \bibinfo{year}{2018}.
\newblock \bibinfo{title}{Modeling cities and landscapes in 3d with citygml},
  in: \bibinfo{booktitle}{Building Information Modeling: Technology Foundations
  and Industry Practice}. \bibinfo{publisher}{Springer International
  Publishing}, \bibinfo{address}{Cham}, pp. \bibinfo{pages}{199--215}.
\newblock \DOIprefix\doi{10.1007/978-3-319-92862-3_11}.
\bibitem[{Besl and McKay(1992)}]{besl1992icp}
\bibinfo{author}{Besl, P.J.}, \bibinfo{author}{McKay, N.D.},
  \bibinfo{year}{1992}.
\newblock \bibinfo{title}{{Method for registration of 3-D shapes}}, in:
  \bibinfo{editor}{Schenker, P.S.} (Ed.), \bibinfo{booktitle}{Sensor Fusion IV:
  Control Paradigms and Data Structures}, \bibinfo{organization}{International
  Society for Optics and Photonics}. \bibinfo{publisher}{SPIE}. pp.
  \bibinfo{pages}{586 -- 606}.
\newblock \URLprefix \url{https://doi.org/10.1117/12.57955},
  \DOIprefix\doi{10.1117/12.57955}.
\bibitem[{Biber and Strasser(2003)}]{biber2003ndt}
\bibinfo{author}{Biber, P.}, \bibinfo{author}{Strasser, W.},
  \bibinfo{year}{2003}.
\newblock \bibinfo{title}{The normal distributions transform: a new approach to
  laser scan matching}, in: \bibinfo{booktitle}{Proceedings 2003 IEEE/RSJ
  International Conference on Intelligent Robots and Systems (IROS 2003) (Cat.
  No.03CH37453)}, pp. \bibinfo{pages}{2743--2748 vol.3}.
\newblock \DOIprefix\doi{10.1109/IROS.2003.1249285}.
\bibitem[{Chebrolu et~al.(2021)Chebrolu, Läbe, Vysotska, Behley and
  Stachniss}]{bouguelia2018robustkernel}
\bibinfo{author}{Chebrolu, N.}, \bibinfo{author}{Läbe, T.},
  \bibinfo{author}{Vysotska, O.}, \bibinfo{author}{Behley, J.},
  \bibinfo{author}{Stachniss, C.}, \bibinfo{year}{2021}.
\newblock \bibinfo{title}{Adaptive robust kernels for non-linear least squares
  problems}.
\newblock \bibinfo{journal}{IEEE Robotics and Automation Letters}
  \bibinfo{volume}{6}, \bibinfo{pages}{2240--2247}.
\newblock \DOIprefix\doi{10.1109/LRA.2021.3061331}.
\bibitem[{Chen and Medioni(1992)}]{PointtoPlane}
\bibinfo{author}{Chen, Y.}, \bibinfo{author}{Medioni, G.},
  \bibinfo{year}{1992}.
\newblock \bibinfo{title}{Object modelling by registration of multiple range
  images}.
\newblock \bibinfo{journal}{Image and Vision Computing} \bibinfo{volume}{10},
  \bibinfo{pages}{145--155}.
\newblock \URLprefix
  \url{https://www.sciencedirect.com/science/article/pii/026288569290066C},
  \DOIprefix\doi{https://doi.org/10.1016/0262-8856(92)90066-C}.
  \bibinfo{note}{range Image Understanding}.
\bibitem[{Choy et~al.(2020)Choy, Dong and Koltun}]{choy2020dgr}
\bibinfo{author}{Choy, C.}, \bibinfo{author}{Dong, W.},
  \bibinfo{author}{Koltun, V.}, \bibinfo{year}{2020}.
\newblock \bibinfo{title}{Deep global registration}, in:
  \bibinfo{booktitle}{2020 IEEE/CVF Conference on Computer Vision and Pattern
  Recognition (CVPR)}, pp. \bibinfo{pages}{2511--2520}.
\newblock \DOIprefix\doi{10.1109/CVPR42600.2020.00259}.
\bibitem[{Eriksson and Harrie(2021)}]{Eriksson2021}
\bibinfo{author}{Eriksson, H.}, \bibinfo{author}{Harrie, L.},
  \bibinfo{year}{2021}.
\newblock \bibinfo{title}{Versioning of 3d city models for municipality
  applications: Needs, obstacles and recommendations}.
\newblock \bibinfo{journal}{ISPRS International Journal of Geo-Information}
  \bibinfo{volume}{10}.
\newblock \URLprefix \url{https://www.mdpi.com/2220-9964/10/2/55},
  \DOIprefix\doi{10.3390/ijgi10020055}.
\bibitem[{Fischler and Bolles(1981)}]{fischler1981ransac}
\bibinfo{author}{Fischler, M.A.}, \bibinfo{author}{Bolles, R.C.},
  \bibinfo{year}{1981}.
\newblock \bibinfo{title}{Random sample consensus: a paradigm for model fitting
  with applications to image analysis and automated cartography}.
\newblock \bibinfo{journal}{Commun. ACM} \bibinfo{volume}{24},
  \bibinfo{pages}{381–395}.
\newblock \URLprefix \url{https://doi.org/10.1145/358669.358692},
  \DOIprefix\doi{10.1145/358669.358692}.
\bibitem[{Gelfand et~al.(2003)Gelfand, Ikemoto, Rusinkiewicz and
  Levoy}]{FrameFail}
\bibinfo{author}{Gelfand, N.}, \bibinfo{author}{Ikemoto, L.},
  \bibinfo{author}{Rusinkiewicz, S.}, \bibinfo{author}{Levoy, M.},
  \bibinfo{year}{2003}.
\newblock \bibinfo{title}{Geometrically stable sampling for the icp algorithm},
  in: \bibinfo{booktitle}{Fourth International Conference on 3-D Digital
  Imaging and Modeling, 2003. 3DIM 2003. Proceedings.}, pp.
  \bibinfo{pages}{260--267}.
\newblock \DOIprefix\doi{10.1109/IM.2003.1240258}.
\bibitem[{Hess et~al.(2016)Hess, Kohler, Rapp and Andor}]{Slam2}
\bibinfo{author}{Hess, W.}, \bibinfo{author}{Kohler, D.},
  \bibinfo{author}{Rapp, H.}, \bibinfo{author}{Andor, D.},
  \bibinfo{year}{2016}.
\newblock \bibinfo{title}{Real-time loop closure in 2d lidar slam}, in:
  \bibinfo{booktitle}{2016 IEEE International Conference on Robotics and
  Automation (ICRA)}, pp. \bibinfo{pages}{1271--1278}.
\newblock \DOIprefix\doi{10.1109/ICRA.2016.7487258}.
\bibitem[{Hu et~al.(2020)Hu, Yang, Xie, Rosa, Guo, Wang, Trigoni and
  Markham}]{RandLA}
\bibinfo{author}{Hu, Q.}, \bibinfo{author}{Yang, B.}, \bibinfo{author}{Xie,
  L.}, \bibinfo{author}{Rosa, S.}, \bibinfo{author}{Guo, Y.},
  \bibinfo{author}{Wang, Z.}, \bibinfo{author}{Trigoni, N.},
  \bibinfo{author}{Markham, A.}, \bibinfo{year}{2020}.
\newblock \bibinfo{title}{Randla-net: Efficient semantic segmentation of
  large-scale point clouds}, in: \bibinfo{booktitle}{2020 IEEE/CVF Conference
  on Computer Vision and Pattern Recognition (CVPR)}, pp.
  \bibinfo{pages}{11105--11114}.
\newblock \DOIprefix\doi{10.1109/CVPR42600.2020.01112}.
\bibitem[{Koszyk et~al.(2024)Koszyk, Jasi\'{n}ska, Pargie\l{}a, Malczewska,
  Grzelka, Bieda and Ambrozi\'{n}ski}]{koszyk2024segment}
\bibinfo{author}{Koszyk, J.}, \bibinfo{author}{Jasi\'{n}ska, A.},
  \bibinfo{author}{Pargie\l{}a, K.}, \bibinfo{author}{Malczewska, A.},
  \bibinfo{author}{Grzelka, K.}, \bibinfo{author}{Bieda, A.},
  \bibinfo{author}{Ambrozi\'{n}ski, {\L}.}, \bibinfo{year}{2024}.
\newblock \bibinfo{title}{Semantic segmentation-driven integration of point
  clouds from mobile scanning platforms in urban environments}.
\newblock \bibinfo{journal}{Remote Sensing} \bibinfo{volume}{16},
  \bibinfo{pages}{3434}.
\newblock \DOIprefix\doi{10.3390/rs16183434}.
\bibitem[{Lague et~al.(2013)Lague, Brodu and Leroux}]{M3C2}
\bibinfo{author}{Lague, D.}, \bibinfo{author}{Brodu, N.},
  \bibinfo{author}{Leroux, J.}, \bibinfo{year}{2013}.
\newblock \bibinfo{title}{Accurate 3d comparison of complex topography with
  terrestrial laser scanner: Application to the rangitikei canyon (n-z)}.
\newblock \bibinfo{journal}{ISPRS Journal of Photogrammetry and Remote Sensing}
  \bibinfo{volume}{82}, \bibinfo{pages}{10--26}.
\newblock \URLprefix
  \url{https://www.sciencedirect.com/science/article/pii/S0924271613001184},
  \DOIprefix\doi{https://doi.org/10.1016/j.isprsjprs.2013.04.009}.
\bibitem[{Lan et~al.(2019)Lan, Yew and Lee}]{zhu2020robust}
\bibinfo{author}{Lan, Z.}, \bibinfo{author}{Yew, Z.J.}, \bibinfo{author}{Lee,
  G.H.}, \bibinfo{year}{2019}.
\newblock \bibinfo{title}{Robust point cloud based reconstruction of
  large-scale outdoor scenes}, in: \bibinfo{booktitle}{2019 IEEE/CVF Conference
  on Computer Vision and Pattern Recognition (CVPR)}, pp.
  \bibinfo{pages}{9682--9690}.
\newblock \DOIprefix\doi{10.1109/CVPR.2019.00992}.
\bibitem[{Lee et~al.(2024)Lee, Kwon, Kim, Choi and Sohn}]{Lee2024}
\bibinfo{author}{Lee, E.}, \bibinfo{author}{Kwon, Y.}, \bibinfo{author}{Kim,
  C.}, \bibinfo{author}{Choi, W.}, \bibinfo{author}{Sohn, H.G.},
  \bibinfo{year}{2024}.
\newblock \bibinfo{title}{Multi-source point cloud registration for urban areas
  using a coarse-to-fine approach}.
\newblock \bibinfo{journal}{GIScience \& Remote Sensing} \bibinfo{volume}{61},
  \bibinfo{pages}{2341557}.
\newblock \DOIprefix\doi{10.1080/15481603.2024.2341557}.
\bibitem[{Lee and Jung(2021)}]{lee2021curvseg}
\bibinfo{author}{Lee, H.}, \bibinfo{author}{Jung, J.}, \bibinfo{year}{2021}.
\newblock \bibinfo{title}{Clustering-based plane segmentation neural network
  for urban scene modeling}.
\newblock \bibinfo{journal}{Sensors} \bibinfo{volume}{21}.
\newblock \URLprefix \url{https://www.mdpi.com/1424-8220/21/24/8382},
  \DOIprefix\doi{10.3390/s21248382}.
\bibitem[{Lucks et~al.(2021)Lucks, Klingbeil, Plümer and
  Dehbi}]{lucks2021improving}
\bibinfo{author}{Lucks, L.}, \bibinfo{author}{Klingbeil, L.},
  \bibinfo{author}{Plümer, L.}, \bibinfo{author}{Dehbi, Y.},
  \bibinfo{year}{2021}.
\newblock \bibinfo{title}{Improving trajectory estimation using 3d city models
  and kinematic point clouds}.
\newblock \bibinfo{journal}{Transactions in GIS} \bibinfo{volume}{25},
  \bibinfo{pages}{238--260}.
\newblock \DOIprefix\doi{10.1111/tgis.12719}.
\bibitem[{Magnusson(2009)}]{magnusson2009ndt}
\bibinfo{author}{Magnusson, M.}, \bibinfo{year}{2009}.
\newblock \bibinfo{title}{The Three-Dimensional Normal-Distributions Transform:
  An Efficient Representation for Registration, Surface Analysis and Loop
  Detection}.
\newblock Ph.D. thesis. Örebro University.
\bibitem[{Mahmood et~al.(2020)Mahmood, Han and Lee}]{mahmood2020crosssec}
\bibinfo{author}{Mahmood, B.}, \bibinfo{author}{Han, S.}, \bibinfo{author}{Lee,
  D.E.}, \bibinfo{year}{2020}.
\newblock \bibinfo{title}{Bim-based registration and localization of 3d point
  clouds of indoor scenes using geometric features for augmented reality}.
\newblock \bibinfo{journal}{Remote Sensing} \bibinfo{volume}{12}.
\newblock \URLprefix \url{https://www.mdpi.com/2072-4292/12/14/2302},
  \DOIprefix\doi{10.3390/rs12142302}.
\bibitem[{Mellado et~al.(2014)Mellado, Aiger and Mitra}]{mellado2014super4pcs}
\bibinfo{author}{Mellado, N.}, \bibinfo{author}{Aiger, D.},
  \bibinfo{author}{Mitra, N.J.}, \bibinfo{year}{2014}.
\newblock \bibinfo{title}{Super 4pcs fast global pointcloud registration via
  smart indexing}.
\newblock \bibinfo{journal}{Computer Graphics Forum} \bibinfo{volume}{33},
  \bibinfo{pages}{205--215}.
\newblock \DOIprefix\doi{https://doi.org/10.1111/cgf.12446}.
\bibitem[{Myronenko and Song(2010)}]{myronenko2010cpd}
\bibinfo{author}{Myronenko, A.}, \bibinfo{author}{Song, X.},
  \bibinfo{year}{2010}.
\newblock \bibinfo{title}{Point set registration: Coherent point drift}.
\newblock \bibinfo{journal}{IEEE Transactions on Pattern Analysis and Machine
  Intelligence} \bibinfo{volume}{32}, \bibinfo{pages}{2262--2275}.
\newblock \DOIprefix\doi{10.1109/TPAMI.2010.46}.
\bibitem[{Qin et~al.(2024)Qin, Zhou, Liu, Zhang, Cheng and Guo}]{Liu2022}
\bibinfo{author}{Qin, H.}, \bibinfo{author}{Zhou, Y.}, \bibinfo{author}{Liu,
  C.}, \bibinfo{author}{Zhang, X.}, \bibinfo{author}{Cheng, Z.},
  \bibinfo{author}{Guo, J.}, \bibinfo{year}{2024}.
\newblock \bibinfo{title}{Sarnet: Semantic augmented registration
  of large-scale urban point clouds}, in: \bibinfo{editor}{Zhang, F.L.},
  \bibinfo{editor}{Sharf, A.} (Eds.), \bibinfo{booktitle}{Computational Visual
  Media}, \bibinfo{publisher}{Springer Nature Singapore},
  \bibinfo{address}{Singapore}. pp. \bibinfo{pages}{152--174}.
\bibitem[{Raffl and Holst(2022)}]{lucksRaffl}
\bibinfo{author}{Raffl, L.}, \bibinfo{author}{Holst, C.}, \bibinfo{year}{2022}.
\newblock \bibinfo{title}{Including virtual target points from laser scanning
  into the point-wise rigorous deformation analysis at geo-monitoring
  applications}.
\newblock \bibinfo{journal}{5th Joint International Symposium on Deformation
  Monitoring (JISDM)} , \bibinfo{pages}{20.–22.06.2022}.
\bibitem[{Rusu et~al.(2009)Rusu, Blodow and Beetz}]{rusu2009fpfh}
\bibinfo{author}{Rusu, R.B.}, \bibinfo{author}{Blodow, N.},
  \bibinfo{author}{Beetz, M.}, \bibinfo{year}{2009}.
\newblock \bibinfo{title}{Fast point feature histograms (fpfh) for 3d
  registration}, in: \bibinfo{booktitle}{2009 IEEE International Conference on
  Robotics and Automation}, pp. \bibinfo{pages}{3212--3217}.
\newblock \DOIprefix\doi{10.1109/ROBOT.2009.5152473}.
\bibitem[{Rusu and Cousins(2011)}]{rusu2011pcl}
\bibinfo{author}{Rusu, R.B.}, \bibinfo{author}{Cousins, S.},
  \bibinfo{year}{2011}.
\newblock \bibinfo{title}{3d is here: Point cloud library (pcl)}, in:
  \bibinfo{booktitle}{2011 IEEE International Conference on Robotics and
  Automation}, pp. \bibinfo{pages}{1--4}.
\newblock \DOIprefix\doi{10.1109/ICRA.2011.5980567}.
\bibitem[{Segal et~al.(2009)Segal, Haehnel and Thrun}]{segal2009gicp}
\bibinfo{author}{Segal, A.}, \bibinfo{author}{Haehnel, D.},
  \bibinfo{author}{Thrun, S.}, \bibinfo{year}{2009}.
\newblock \bibinfo{title}{Generalized-icp.}, in: \bibinfo{booktitle}{Robotics:
  science and systems}, \bibinfo{organization}{Seattle, WA}. p.
  \bibinfo{pages}{435}.
\bibitem[{Shan et~al.(2020)Shan, Englot, Meyers, Wang, Ratti and Rus}]{Slam1}
\bibinfo{author}{Shan, T.}, \bibinfo{author}{Englot, B.},
  \bibinfo{author}{Meyers, D.}, \bibinfo{author}{Wang, W.},
  \bibinfo{author}{Ratti, C.}, \bibinfo{author}{Rus, D.}, \bibinfo{year}{2020}.
\newblock \bibinfo{title}{Lio-sam: Tightly-coupled lidar inertial odometry via
  smoothing and mapping}, in: \bibinfo{booktitle}{2020 IEEE/RSJ International
  Conference on Intelligent Robots and Systems (IROS)}, pp.
  \bibinfo{pages}{5135--5142}.
\newblock \DOIprefix\doi{10.1109/IROS45743.2020.9341176}.
\bibitem[{Sun et~al.(2023)Sun, Zhong, Wu and Guo}]{sun2023stripadjust}
\bibinfo{author}{Sun, Z.}, \bibinfo{author}{Zhong, R.}, \bibinfo{author}{Wu,
  Q.}, \bibinfo{author}{Guo, J.}, \bibinfo{year}{2023}.
\newblock \bibinfo{title}{Airborne lidar strip adjustment method based on point
  clouds with planar neighborhoods}.
\newblock \bibinfo{journal}{Remote Sensing} \bibinfo{volume}{15}.
\newblock \URLprefix \url{https://www.mdpi.com/2072-4292/15/23/5447},
  \DOIprefix\doi{10.3390/rs15235447}.
\bibitem[{{Viametris}(2025)}]{ViametrisMS96}
\bibinfo{author}{{Viametris}}, \bibinfo{year}{2025}.
\newblock \bibinfo{title}{Ms-96: Versatile mobile mapping system}.
\newblock \URLprefix \url{https://viametris.com/ms-96/}.
  \bibinfo{note}{accessed: 2025-03-14}.
\bibitem[{Wang et~al.(2020)Wang, Wen, Dai, Yu and Liu}]{Wang2020}
\bibinfo{author}{Wang, C.}, \bibinfo{author}{Wen, C.}, \bibinfo{author}{Dai,
  Y.}, \bibinfo{author}{Yu, S.}, \bibinfo{author}{Liu, M.},
  \bibinfo{year}{2020}.
\newblock \bibinfo{title}{Urban 3d modeling using mobile laser scanning: a
  review}.
\newblock \bibinfo{journal}{Virtual Reality \& Intelligent Hardware}
  \bibinfo{volume}{2}, \bibinfo{pages}{175--212}.
\newblock \URLprefix
  \url{https://www.sciencedirect.com/science/article/pii/S2096579620300395},
  \DOIprefix\doi{https://doi.org/10.1016/j.vrih.2020.05.003}. \bibinfo{note}{3D
  Visual Processing and Reconstruction Special Issue}.
\bibitem[{Xu et~al.(2024)Xu, Han, Zhong, Sang and Zhang}]{xu2024precisereg}
\bibinfo{author}{Xu, M.}, \bibinfo{author}{Han, Y.}, \bibinfo{author}{Zhong,
  X.}, \bibinfo{author}{Sang, F.}, \bibinfo{author}{Zhang, Y.},
  \bibinfo{year}{2024}.
\newblock \bibinfo{title}{A precise registration method for large-scale urban
  point clouds based on phased and spatial geometric features}.
\newblock \bibinfo{journal}{Measurement Science and Technology}
  \bibinfo{volume}{36}, \bibinfo{pages}{015202}.
\newblock \DOIprefix\doi{10.1088/1361-6501/ad7e44}.
\bibitem[{Xu et~al.(2020)Xu, Wang, Wang and Zheng}]{jiang2017nvcluster}
\bibinfo{author}{Xu, S.}, \bibinfo{author}{Wang, R.}, \bibinfo{author}{Wang,
  H.}, \bibinfo{author}{Zheng, H.}, \bibinfo{year}{2020}.
\newblock \bibinfo{title}{An optimal hierarchical clustering approach to mobile
  lidar point clouds}.
\newblock \bibinfo{journal}{IEEE Transactions on Intelligent Transportation
  Systems} \bibinfo{volume}{21}, \bibinfo{pages}{2765--2776}.
\newblock \DOIprefix\doi{10.1109/TITS.2019.2912455}.
\bibitem[{Xu et~al.(2025)Xu, Hackl and Holst}]{XU2025PL4U}
\bibinfo{author}{Xu, Z.}, \bibinfo{author}{Hackl, M.}, \bibinfo{author}{Holst,
  C.}, \bibinfo{year}{2025}.
\newblock \bibinfo{title}{Pl4u: Automated plane-based uncertainty evaluation
  and reduction method for indoor mobile laser scanning systems}.
\newblock \bibinfo{journal}{ISPRS Journal of Photogrammetry and Remote Sensing}
  \bibinfo{volume}{228}, \bibinfo{pages}{467--488}.
\newblock \URLprefix
  \url{https://www.sciencedirect.com/science/article/pii/S0924271625002813},
  \DOIprefix\doi{10.1016/j.isprsjprs.2025.07.019}.
\bibitem[{Yan et~al.(2023)Yan, Wei, Xie, Dai, Wu and Huang}]{yan2022gror}
\bibinfo{author}{Yan, L.}, \bibinfo{author}{Wei, P.}, \bibinfo{author}{Xie,
  H.}, \bibinfo{author}{Dai, J.}, \bibinfo{author}{Wu, H.},
  \bibinfo{author}{Huang, M.}, \bibinfo{year}{2023}.
\newblock \bibinfo{title}{A new outlier removal strategy based on reliability
  of correspondence graph for fast point cloud registration}.
\newblock \bibinfo{journal}{IEEE Transactions on Pattern Analysis and Machine
  Intelligence} \bibinfo{volume}{45}, \bibinfo{pages}{7986--8002}.
\newblock \DOIprefix\doi{10.1109/TPAMI.2022.3226498}.
\bibitem[{Yang et~al.(2021)Yang, Shi and Carlone}]{yang2020teaserpp}
\bibinfo{author}{Yang, H.}, \bibinfo{author}{Shi, J.},
  \bibinfo{author}{Carlone, L.}, \bibinfo{year}{2021}.
\newblock \bibinfo{title}{Teaser: Fast and certifiable point cloud
  registration}.
\newblock \bibinfo{journal}{IEEE Transactions on Robotics}
  \bibinfo{volume}{37}, \bibinfo{pages}{314--333}.
\newblock \DOIprefix\doi{10.1109/TRO.2020.3033695}.
\bibitem[{Yang et~al.(2016)Yang, Li, Campbell and Jia}]{yang2015goicp}
\bibinfo{author}{Yang, J.}, \bibinfo{author}{Li, H.},
  \bibinfo{author}{Campbell, D.}, \bibinfo{author}{Jia, Y.},
  \bibinfo{year}{2016}.
\newblock \bibinfo{title}{Go-icp: A globally optimal solution to 3d icp
  point-set registration}.
\newblock \bibinfo{journal}{IEEE Transactions on Pattern Analysis and Machine
  Intelligence} \bibinfo{volume}{38}, \bibinfo{pages}{2241--2254}.
\newblock \DOIprefix\doi{10.1109/TPAMI.2015.2513405}.
\bibitem[{Yang(2023)}]{yang2023towards}
\bibinfo{author}{Yang, Y.}, \bibinfo{year}{2023}.
\newblock \bibinfo{title}{Towards improved targetless registration and
  deformation analysis of TLS point clouds using patch-based segmentation}.
\newblock Ph.D. thesis. Dissertation, Stuttgart, University of Stuttgart, 2023.
\newblock \DOIprefix\doi{10.18419/opus-13796}.
\bibitem[{Yang and Holst(2025)}]{yang2025piecewise}
\bibinfo{author}{Yang, Y.}, \bibinfo{author}{Holst, C.}, \bibinfo{year}{2025}.
\newblock \bibinfo{title}{Piecewise-icp: Efficient and robust registration for
  4d point clouds in permanent laser scanning}.
\newblock \bibinfo{journal}{ISPRS Journal of Photogrammetry and Remote Sensing}
  \bibinfo{volume}{227}, \bibinfo{pages}{481--500}.
\newblock \DOIprefix\doi{https://doi.org/10.1016/j.isprsjprs.2025.06.026}.
\bibitem[{Yang and Schwieger(2023)}]{YangSchwiegerSVreg}
\bibinfo{author}{Yang, Y.}, \bibinfo{author}{Schwieger, V.},
  \bibinfo{year}{2023}.
\newblock \bibinfo{title}{Supervoxel-based targetless registration and
  identification of stable areas for deformed point clouds}.
\newblock \bibinfo{journal}{Journal of Applied Geodesy} \bibinfo{volume}{17},
  \bibinfo{pages}{161--170}.
\newblock \DOIprefix\doi{10.1515/jag-2022-0031}.
\bibitem[{Zhong(2009)}]{ISS}
\bibinfo{author}{Zhong, Y.}, \bibinfo{year}{2009}.
\newblock \bibinfo{title}{Intrinsic shape signatures: A shape descriptor for 3d
  object recognition}, in: \bibinfo{booktitle}{2009 IEEE 12th International
  Conference on Computer Vision Workshops, ICCV Workshops}, pp.
  \bibinfo{pages}{689--696}.
\newblock \DOIprefix\doi{10.1109/ICCVW.2009.5457637}.

\end{thebibliography}

\end{document}